\documentclass
[
   oneside,
    openright,               
    cleardoublepage = empty, 
    fontsize = 12 pt,        
    american,                
    captions = tableheading, 
    numbers = noenddot,                       
    footheight = 35 pt,      
]
{scrbook}

\newif\ifprintVersion   
\newif\ifprofessionalPrint 
\newif\iffancyTheorems  
\newif\ifboldNumberSets 
\newif\ifbachelorThesis 

\printVersionfalse
\professionalPrintfalse
\fancyTheoremstrue
\boldNumberSetstrue
\bachelorThesistrue

\newcommand*{\printTitle}{}
\newcommand*{\printGermanTitle}{}
\newcommand*{\myTitle}[2]{\renewcommand*{\printTitle}{#1}\renewcommand*{\printGermanTitle}{#2}}
\newcommand*{\printTitleBold}{\textbf{\printTitle}}

\newcommand*{\printAuthor}{}
\newcommand*{\myName}[1]{\renewcommand*{\printAuthor}{#1}}

\newcommand*{\printProgram}{}
\newcommand*{\myProgram}[1]{\renewcommand*{\printProgram}{#1}}

\newcommand*{\printDateReceived}{}
\newcommand*{\dateOfHandingIn}[1]{\renewcommand*{\printDateReceived}{#1}}

\newcommand*{\printSubject}{}
\newcommand*{\mySubject}[1]{\renewcommand*{\printSubject}{#1}}

\newcommand*{\printKeywords}{}
\newcommand*{\myKeywords}[1]{\renewcommand*{\printKeywords}{#1}}

\newcommand*{\printNameOfSupervisor}{}
\newcommand*{\nameOfMySupervisor}[1]{\renewcommand*{\printNameOfSupervisor}{#1}}

\newcommand*{\printAdditionalExaminers}{}
\newcommand*{\additionalExaminers}[1]{\renewcommand*{\printAdditionalExaminers}{#1}}

\newlength{\extraborderlength}
\newcommand*{\extraBorder}[1]{\setlength{\extraborderlength}{#1}}

\newlength{\mybindingcorrection}
\newcommand*{\bindingCorrection}[1]{\setlength{\mybindingcorrection}{#1}} 

    \extraBorder{3 mm}

    \bindingCorrection{6 mm}

    \myTitle{Comparison of Outlier Detection Algorithms on String Data}{Vergleich von Ausreißererkennungsalgorithmen auf Zeichenketten}

    \myName{Philip Maus}

    \myProgram{IT-Systems Engineering}

    \dateOfHandingIn{31. August 2025}

    \nameOfMySupervisor{Prof.\,Dr. Tobias Friedrich}

    \additionalExaminers{Dr. Timo Kötzing\newline Xiaoyue Li}

    \mySubject{A bachelor's thesis comparing the local outlier factor algorithm against a new regular expression learner-based syntactical outlier detection algorithm for single-word string data}

    \myKeywords{bachelor thesis | outlier | detection | classification | labeling | string | HiLRE | LOF | algorithm}

\usepackage[utf8]{inputenc} 
\usepackage[T1]{fontenc}    
\usepackage
[
    ngerman,         
    main = american, 
]
{babel}                     
\usepackage{forest}

\input glyphtounicode
\usepackage{calc} 

\newlength{\myparindent}
\newlength{\myparskip}
\setfootnoterule{0 cm}

\deffootnote[1.2 em]{1.2 em}{0 em}{\makebox[1.4 em][l]{\textbf{\thefootnotemark}}}

\makeatletter%
    \@removefromreset{footnote}{chapter}%
\makeatother

\usepackage[dvipsnames]{xcolor} 

\definecolor{stroke1}{HTML}{2574A9} 

\colorlet{captionlabel}{black}
\colorlet{footerpagenr}{black}
\colorlet{footerchapter}{stroke1}
\colorlet{footerchaptername}{black}
\colorlet{footersection}{stroke1}
\colorlet{footersectionname}{black}
\colorlet{chapternumber}{stroke1}

\newlength{\mypaperwidth}
\newlength{\mypaperheight}
\newlength{\mybodywidth}
\newlength{\mybodyheight}
\newlength{\myoutermargin}
\ifprintVersion
    \ifprofessionalPrint
    \else
    \fi
\else
\fi

\newlength{\mytopmargin}
\ifprintVersion
    \ifprofessionalPrint
    \fi
\fi

\newlength{\myinnermargin}
\ifprintVersion
    \ifprofessionalPrint
    \fi
\fi

\newlength{\mybottommargin}
\ifprintVersion
    \ifprofessionalPrint
    \fi
\fi

\newcommand{\goldenratio}{1.618}

\newlength{\myheadsep} 
\ifprintVersion
    \ifprofessionalPrint
    \fi
\fi

\newlength{\myfootskip} 
\ifprintVersion
    \ifprofessionalPrint
    \fi
\fi

\newlength{\mymargininnersep} 
\newlength{\mymarginoutersep} 
\ifprintVersion
    \ifprofessionalPrint
    \fi
\fi

\newlength{\mymarginwidth} 
\newlength{\mymarginwidthwithinnersep} 
\usepackage
[
    \ifprintVersion
        \ifprofessionalPrint
            paperwidth = \mypaperwidth + 2\extraborderlength + \mybindingcorrection,
            paperheight = \mypaperheight + 2\extraborderlength,
        \else
            paperwidth = \mypaperwidth,
            paperheight = \mypaperheight,
        \fi
    \else
        paperwidth = \mypaperwidth,
        paperheight = \mypaperheight,
    \fi
    textwidth = \mybodywidth,
    textheight = \mybodyheight,
    outer = \myoutermargin,
    top = \mytopmargin,
    headsep = \myheadsep,
    footskip = \myfootskip,
    marginparsep = \mymargininnersep,
    marginparwidth = \mymarginwidth,
]
{geometry} 

\usepackage
[
]
{scrlayer-scrpage} 

\clearpairofpagestyles

\KOMAoptions
{%
    headwidth = \textwidth + \mymarginwidthwithinnersep,%
    footwidth = \myoutermargin : \textwidth,%
}

\automark[chapter]{chapter}
\automark*[section]{}

\lehead%
{%
    \begin{minipage}[b]{\mymarginwidth}%
        \small\raggedleft\normalfont\textsf{\textbf{\color{footerchapter}\chaptername\ \thechapter}}
    \end{minipage}
}
\cehead{\hspace*{\mymarginwidthwithinnersep}\parbox{\textwidth}{\raggedright\leftmark}}

\rohead%
{%
    \Ifstr{\rightmark}{\leftmark}%
    {%
        \begin{minipage}[b]{\mymarginwidth}%
            \small\raggedright\normalfont\textsf{\textbf{\color{footersection}Chapter\ \thechapter}}%
        \end{minipage}%
    }%
    {%
        \begin{minipage}[b]{\mymarginwidth}%
            \small\raggedright\normalfont\textsf{\textbf{\color{footersection}Section\ \thesection}}%
        \end{minipage}%
    }%
}
\cohead{\hspace*{-\mymarginwidthwithinnersep}\parbox{\textwidth}{\raggedleft\rightmark}}

\lefoot*%
{%
    \vspace*{1 ex}%
    {\color{stroke1}\rule{\myoutermargin - \mymargininnersep}{0.5 mm}}\\
    \begin{minipage}[b]{\myoutermargin - \mymargininnersep}%
        \raggedleft\normalfont\color{footerpagenr}\textbf{\thepage}%
    \end{minipage}%
}
\rofoot*%
{%
    {\color{stroke1}\rule{\myoutermargin - \mymargininnersep}{0.5 mm}}\\
    \begin{minipage}[b]{\myoutermargin - \mymargininnersep}%
        \raggedright\normalfont\color{footerpagenr}\textbf{\thepage}%
    \end{minipage}%
}

\usepackage{caption}
\usepackage{tikz} 

\newlength{\mytmpa}
\newlength{\mytmpb}

\renewcommand*{\partlineswithprefixformat}[3]%
{%
    #2
    \thispagestyle{empty}
    \ifprintVersion
        \setlength{\mytmpa}{0.618\mypaperwidth + \mybindingcorrection + \extraborderlength}%
        \setlength{\mytmpb}{0.382\mypaperheight + \extraborderlength}%
    \else
        \setlength{\mytmpa}{0.618\mypaperwidth}%
        \setlength{\mytmpb}{0.382\mypaperheight}%
    \fi
    \begin{tikzpicture}[overlay, remember picture]%
        \node [inner sep = 0, outer sep = 0, anchor = north] at (current page.north west)%
        {%
            \begin{tikzpicture}[overlay, remember picture]%
            \draw[color = stroke1, line width = 0.7 mm] (\mytmpa, 0) -- (\mytmpa, -\mytmpb);%
            \end{tikzpicture}%
        };%
        \node (align) [align = right, below = \mytmpb - 2 ex, inner sep = 0, outer sep = 0, anchor = north west] at (current page.north west)%
        {%
            \hspace{\mytmpa}\hspace{0.5 em}\partname\ \thepart\\[1 ex]
            \color{stroke1}#3%
        };%
    \end{tikzpicture}%
}
\RedeclareSectionCommand%
[%
    font = \normalfont\Huge\sffamily,
    prefixfont = \normalfont\Huge\sffamily,
]
{part}

\usepackage{etoolbox}
\usepackage[strict]{changepage}
\usetikzlibrary{calc}

\newbool{chapterHasANumber}
\newbool{chapterHasAStar}
\newcommand*{\thischapterlabel}{}
\renewcommand*{\chapterlinesformat}[3]%
{%
    \Ifnumbered{#1}{\setbool{chapterHasANumber}{true}}{\setbool{chapterHasANumber}{false}}%
    \Ifstr{#2}{}{\setbool{chapterHasAStar}{true}}{\setbool{chapterHasAStar}{false}}%
    \ifboolexpr{bool{chapterHasANumber} and not bool{chapterHasAStar}}%
    {%
        \renewcommand*{\thischapterlabel}{\color{chapternumber}\thechapter}
    }%
    {%
        \renewcommand*{\thischapterlabel}{\vphantom{0}}
    }%
    \checkoddpage
    \begin{tikzpicture}[overlay, remember picture]%
    \ifoddpage%
        \coordinate (anchor) at ($(current page.north west) + (\myinnermargin, -\mytopmargin)$);
        \node [inner sep = 0, outer sep = 0, anchor = north west] (numbernode) at (anchor)%
        {%
            \sffamily\fontsize{60}{60}\selectfont\thischapterlabel
        };%
        \draw[color = stroke1, line width = 0.7 mm] ($(numbernode.south west)-(0, 1 ex)$) -- ++(\paperwidth, 0);%
        \ifboolexpr{bool{chapterHasANumber} and not bool{chapterHasAStar}}%
        {%
            \node (align) [text width = \textwidth - 2 cm, align = right, right = \mybodywidth, inner sep = 0, outer sep = 0, anchor = east] at (numbernode.west) {#3};
        }%
        {%
            \node (align) [align = left, inner sep = 0, outer sep = 0, anchor = west] at (numbernode.west) {#3};
        }%
    \else%
        \coordinate (anchor) at ($(current page.north east) - (\myinnermargin, \mytopmargin)$);
        \node [inner sep = 0, outer sep = 0, anchor = north east] (numbernode) at (anchor)%
        {%
            \sffamily\fontsize{60}{60}\selectfont\thischapterlabel
        };%
        \draw[color = stroke1, line width = 0.7 mm] ($(numbernode.south east)-(0, 1 ex)$) -- ++(-\paperwidth, 0);%
        \ifboolexpr{bool{chapterHasANumber} and not bool{chapterHasAStar}}%
        {%
            \node (align) [text width = \textwidth - 2 cm, align = left, left = \mybodywidth, inner sep = 0, outer sep = 0, anchor = west] at (numbernode.east) {#3};
        }%
        {%
            \node (align) [align = right, inner sep = 0, outer sep = 0, anchor = east] at (numbernode.east) {#3};
        }%
    \fi%
    \end{tikzpicture}%
}
\RedeclareSectionCommand%
[%
    font = \color{stroke1}\normalfont\huge\sffamily,
    afterskip = 20 pt,
]
{chapter}

\BeforeStartingTOC[toc]{\pagestyle{plain}}
\AfterStartingTOC{\thispagestyle{plain}}
\usepackage
[
    sortcites,              
    style = alphabetic,     
    defernumbers,           
    safeinputenc,           
    backref = true,         
    backrefstyle = three,   
    hyperref = true,        
    maxbibnames = 99,       
    maxcitenames = 2,       
]
{biblatex} 

\renewbibmacro{in:}%
{%
    \ifentrytype{article}{}{\printtext{\bibstring{in}\intitlepunct}}%
}

\renewbibmacro*{volume+number+eid}%
{%
    \printfield{volume}%
    \iffieldundef{number}{}{\addcolon}%
    \printfield{number}%
    \setunit*{\addcomma\space}%
    \printfield{eid}%
}

\DefineBibliographyStrings{english}%
{%
    backrefpage  = {\lowercase{s}ee page}, 
    backrefpages = {\lowercase{s}ee pages} 
}

\DeclareFieldFormat[article]{title}{\textbf{\color{stroke1}#1}}
\DeclareFieldFormat[inproceedings]{title}{\textbf{\color{stroke1}#1}}
\DeclareFieldFormat[thesis]{title}{\textbf{\color{stroke1}#1}}
\DeclareFieldFormat[book]{title}{\textbf{\color{stroke1}#1}}
\DeclareFieldFormat[unpublished]{title}{\textbf{\color{stroke1}#1}}
\DeclareFieldFormat[report]{title}{\textbf{\color{stroke1}#1}}
\DeclareFieldFormat[inbook]{chapter}{\textbf{\color{stroke1}#1}}
\DeclareFieldFormat[inbook]{title}{#1}
\DeclareFieldFormat{pages}{#1}

\newtoggle{authorend}
\togglefalse{authorend}

\DeclareBibliographyDriver{article}%
{%
  \usebibmacro{bibindex}%
  \usebibmacro{begentry}%
  \iftoggle{authorend}{}{\usebibmacro{author/translator+others}}%
  \setunit{\labelnamepunct}\newblock
  \usebibmacro{title}%
  \newunit
  \printlist{language}%
  \newunit\newblock
  \usebibmacro{byauthor}%
  \newunit\newblock
  \usebibmacro{bytranslator+others}%
  \newunit\newblock
  \printfield{version}%
  \newunit\newblock
  \usebibmacro{in:}%
  \usebibmacro{journal+issuetitle}%
  \newunit
  \usebibmacro{byeditor+others}%
  \newunit
  \usebibmacro{note+pages}%
  \newunit\newblock
  \iftoggle{bbx:isbn}
  {\printfield{issn}}
  {}%
  \newunit\newblock
  \usebibmacro{doi+eprint+url}%
  \newunit\newblock
  \usebibmacro{addendum+pubstate}%
  \setunit{\bibpagerefpunct}\newblock
  \usebibmacro{pageref}%
  \newunit\newblock
  \iftoggle{bbx:related}
  {\usebibmacro{related:init}%
    \usebibmacro{related}}
  {}%
  \usebibmacro{finentry}%
  \iftoggle{authorend}{\usebibmacro{author/translator+others}}{}%
}

\DeclareBibliographyDriver{inbook}%
{%
  \usebibmacro{bibindex}%
  \usebibmacro{begentry}%
  \iftoggle{authorend}{}{\usebibmacro{author/translator+others}}%
  \setunit{\labelnamepunct}\newblock
  \usebibmacro{chapter+pages}%
  \newunit
  \printlist{language}%
  \newunit\newblock
  \usebibmacro{byauthor}%
  \newunit\newblock
  \usebibmacro{in:}%
  \usebibmacro{bybookauthor}%
  \newunit\newblock
  \usebibmacro{maintitle+booktitle}%
  \newunit\newblock
  \usebibmacro{byeditor+others}%
  \newunit\newblock
  \printfield{edition}%
  \newunit
  \iffieldundef{maintitle}
  {\printfield{volume}%
    \printfield{part}}
  {}%
  \newunit
  \printfield{volumes}%
  \newunit\newblock
  \usebibmacro{series+number}%
  \newunit\newblock
  \printfield{note}%
  \newunit\newblock
  \usebibmacro{publisher+location+date}%
  \newunit\newblock
  \newunit\newblock
  \iftoggle{bbx:isbn}
  {\printfield{isbn}}
  {}%
  \newunit\newblock
  \usebibmacro{doi+eprint+url}%
  \newunit\newblock
  \usebibmacro{addendum+pubstate}%
  \setunit{\bibpagerefpunct}\newblock
  \usebibmacro{pageref}%
  \newunit\newblock
  \iftoggle{bbx:related}
  {\usebibmacro{related:init}%
    \usebibmacro{related}}
  {}%
  \usebibmacro{finentry}%
  \iftoggle{authorend}{\usebibmacro{author/translator+others}}{}%
}

\DeclareBibliographyDriver{inproceedings}%
{%
  \usebibmacro{bibindex}%
  \usebibmacro{begentry}%
  \iftoggle{authorend}{}{\usebibmacro{author/translator+others}}%
  \setunit{\labelnamepunct}\newblock
  \usebibmacro{title}%
  \newunit
  \printlist{language}%
  \newunit\newblock
  \usebibmacro{byauthor}%
  \newunit\newblock
  \usebibmacro{in:}%
  \usebibmacro{maintitle+booktitle}%
  \newunit\newblock
  \usebibmacro{event+venue+date}%
  \newunit\newblock
  \usebibmacro{byeditor+others}%
  \newunit\newblock
  \iffieldundef{maintitle}
  {\printfield{volume}%
    \printfield{part}}
  {}%
  \newunit
  \printfield{volumes}%
  \newunit\newblock
  \usebibmacro{series+number}%
  \newunit\newblock
  \printfield{note}%
  \newunit\newblock
  \printlist{organization}%
  \newunit
  \usebibmacro{publisher+location+date}%
  \newunit\newblock
  \usebibmacro{chapter+pages}%
  \newunit\newblock
  \iftoggle{bbx:isbn}
  {\printfield{isbn}}
  {}%
  \newunit\newblock
  \usebibmacro{doi+eprint+url}%
  \newunit\newblock
  \usebibmacro{addendum+pubstate}%
  \setunit{\bibpagerefpunct}\newblock
  \usebibmacro{pageref}%
  \newunit\newblock
  \iftoggle{bbx:related}
  {\usebibmacro{related:init}%
    \usebibmacro{related}}
  {}%
  \usebibmacro{finentry}%
  \iftoggle{authorend}{\usebibmacro{author/translator+others}}{}%
}

\DeclareBibliographyDriver{thesis}%
{%
  \usebibmacro{bibindex}%
  \usebibmacro{begentry}%
  \iftoggle{authorend}{}{\usebibmacro{author}}%
  \setunit{\labelnamepunct}\newblock
  \usebibmacro{title}%
  \newunit
  \printlist{language}%
  \newunit\newblock
  \usebibmacro{byauthor}%
  \newunit\newblock
  \printfield{note}%
  \newunit\newblock
  \printfield{type}%
  \newunit
  \usebibmacro{institution+location+date}%
  \newunit\newblock
  \usebibmacro{chapter+pages}%
  \newunit
  \printfield{pagetotal}%
  \newunit\newblock
  \iftoggle{bbx:isbn}
  {\printfield{isbn}}
  {}%
  \newunit\newblock
  \usebibmacro{doi+eprint+url}%
  \newunit\newblock
  \usebibmacro{addendum+pubstate}%
  \setunit{\bibpagerefpunct}\newblock
  \usebibmacro{pageref}%
  \newunit\newblock
  \iftoggle{bbx:related}
  {\usebibmacro{related:init}%
    \usebibmacro{related}}
  {}%
  \usebibmacro{finentry}%
  \iftoggle{authorend}{\usebibmacro{author}}{}%
}

\providebool{bbx:subentry}
\newbibmacro*{citenum}%
{
  \printtext[bibhyperref]{
    \printfield{prefixnumber}%
    \printfield{labelnumber}%
    \ifbool{bbx:subentry}
    {\printfield{entrysetcount}}
    {}}%
}

\DeclareCiteCommand{\conline}[\mkbibbrackets]
{\usebibmacro{prenote}}
{\usebibmacro{citeindex}%
  \usebibmacro{citenum}}
{\multicitedelim}
{\usebibmacro{postnote}}       
\usepackage
[
    babel = true, 
]
{microtype}           
\usepackage{csquotes} 

\usepackage{amsmath}
\usepackage{amssymb}
\usepackage{amsthm}
\usepackage{thmtools}
\usepackage{mathtools}
\usepackage{thm-restate}
\usepackage{dsfont}        
\usepackage{braceMnSymbol} 

\usepackage
[
    ttscale = 0.85, 
]
{libertine} 
\usepackage
[
    libertine,    
    slantedGreek, 
    vvarbb,       
    libaltvw,     
]
{newtxmath} 
\usepackage{url} 
\usepackage{bm}  

\usepackage{graphicx} 
\usepackage
[
    subrefformat = simple, 
    labelformat = simple,  
]
{subcaption}         
\usepackage{wrapfig} 

\columnsep = \mymargininnersep

\usepackage{array}     
\usepackage{booktabs}  
\usepackage{longtable} 
\usepackage{pdflscape} 

\usepackage[shortlabels]{enumitem} 

\usepackage
[
    ruled,         
    vlined,        
    linesnumbered, 
]
{algorithm2e} 

\usepackage{xspace}   
\usepackage
[
    shortcuts, 
]
{extdash}             
\usepackage{setspace} 

\usepackage{xparse}    
\usepackage{footnote}  
\usepackage{afterpage} 
\usepackage
[
    textsize = scriptsize, 
]
{todonotes}            

\usepackage
[
    bookmarks = true,                 
    bookmarksopen = false,            
    bookmarksnumbered = true,         
    pdfstartpage = 1,                 
    pdftitle = {{\printTitle}},       
    pdfauthor = {{\printAuthor}},     
    pdfsubject = {{\printSubject}},   
    pdfkeywords = {{\printKeywords}}, 
    breaklinks = true,                
    \ifprintVersion
        hidelinks,                    
    \else
    colorlinks = true,            
    allcolors = stroke1,          
    \fi
]
{hyperref} 

\usepackage
[
    noabbrev,   
    nameinlink, 
]
{cleveref} 
\newcommand*{\colloquialDegreeName}{Master}
\newcommand*{\colloquialDegreeNameLowercase}{master}

\newcommand*{\degreeAbbreviation}{M.}

\ifbachelorThesis
    \renewcommand*{\colloquialDegreeName}{Bachelor}
    \renewcommand*{\colloquialDegreeNameLowercase}{bachelor}
    \renewcommand*{\degreeAbbreviation}{B.}
\fi

\makeatletter
    \def\IfEmptyTF#1%
    {%
        \if\relax\detokenize{#1}\relax%
            \expandafter\@firstoftwo%
        \else%
            \expandafter\@secondoftwo%
        \fi%
    }
\makeatother

\NewDocumentCommand{\mathOrText}{m}
{%
    \ensuremath{#1}\xspace%
}

\let\originalleft\left
\let\originalright\right
\renewcommand{\left}{\mathopen{}\mathclose\bgroup\originalleft}
\renewcommand{\right}{\aftergroup\egroup\originalright}

\makeatletter
    \DeclareRobustCommand{\bfseries}%
    {%
        \not@math@alphabet\bfseries\mathbf%
        \fontseries\bfdefault\selectfont%
        \boldmath%
    }
\makeatother

\xspaceaddexceptions{]\}}

\allowdisplaybreaks

\crefname{ineq}{inequality}{inequalities}
\creflabelformat{ineq}{#2{\upshape(#1)}#3} 

\crefname{term}{term}{terms}
\creflabelformat{term}{#2{\upshape(#1)}#3}

\let\oldfootnote\footnote

\newlength{\spaceBeforeFootnote} 
\newlength{\spaceAfterFootnote}  

\RenewDocumentCommand{\footnote}{o o o m}%
{%
    \IfNoValueTF{#1}%
    {%
        \oldfootnote{#4}%
    }%
    {%
        \setlength{\spaceBeforeFootnote}{\IfEmptyTF{#1}{0}{#1} em}%
        \IfNoValueTF{#2}%
        {%
            \hspace*{\spaceBeforeFootnote}\oldfootnote{#4}%
        }%
        {%
            \setlength{\spaceAfterFootnote}{\IfEmptyTF{#2}{0}{#2} em}%
            \hspace*{\spaceBeforeFootnote}\IfNoValueTF{#3}{\oldfootnote{#4}}{\oldfootnote[#3]{#4}}\hspace*{\spaceAfterFootnote}%
        }%
    }%
}

\makesavenoteenv{figure}
\makesavenoteenv{table}
\makesavenoteenv{tabular}

\iffancyTheorems
    \declaretheoremstyle
    [
        spaceabove = \topsep,
        spacebelow = \topsep,
        headfont = \bfseries,
        headformat = \textcolor{stroke1}{$\blacktriangleright$} \NAME~\NUMBER \NOTE,
        notefont = \bfseries,
        notebraces = {(}{)},
        bodyfont = \normalfont,
        postheadspace = 0.5 em,
        qed = \textcolor{stroke1}{\bfseries$\blacktriangleleft$},
    ]
    {myTheoremStyle}
    
    \declaretheorem
    [
        style = myTheoremStyle,
        name = Conjecture,
        numberwithin = chapter,
    ]
    {conjecture}
    \declaretheorem
    [
        style = myTheoremStyle,
        name = Proposition,
        sharenumber = conjecture,
    ]
    {proposition}
    \declaretheorem
    [
        style = myTheoremStyle,
        name = Claim,
        sharenumber = conjecture,
    ]
    {claim}
    \declaretheorem
    [
        style = myTheoremStyle,
        name = Lemma,
        sharenumber = conjecture,
    ]
    {lemma}
    \declaretheorem
    [
        style = myTheoremStyle,
        name = Corollary,
        sharenumber = conjecture,
    ]
    {corollary}
    \declaretheorem
    [
        style = myTheoremStyle,
        name = Theorem,
        sharenumber = conjecture,
    ]
    {theorem}
    \declaretheorem
    [
        style = myTheoremStyle,
        name = Definition,
        sharenumber = conjecture,
    ]
    {definition}
    \declaretheorem
    [
        style = myTheoremStyle,
        name = Example,
        sharenumber = conjecture,
    ]
    {example}
    \declaretheorem
    [
        style = myTheoremStyle,
        name = Remark,
        sharenumber = conjecture,
    ]
    {remark}
\else
    \theoremstyle{plain}
    
    \newtheorem{conjecture}{Conjecture}[chapter]
    \newtheorem{proposition}[conjecture]{Proposition}

    \newtheorem{definition}[conjecture]{Definition}
    \newtheorem{example}[conjecture]{Example}
    
\fi

\NewDocumentCommand{\functionTemplate}{m m m m o}%
{%
    \IfNoValueTF{#5}%
    {%
        \mathOrText{#1\left#2{#4}\right#3}%
    }%
    {%
        \mathOrText{#1#5#2{#4}#5#3}%
    }%
}

\newcommand*{\leftBracketType}{(}
\newcommand*{\rightBracketType}{)}

\NewDocumentCommand{\createFunction}{m m o o}%
{%
    \renewcommand*{\leftBracketType}{\IfNoValueTF{#3}{(}{#3}}%
    \renewcommand*{\rightBracketType}{\IfNoValueTF{#4}{)}{#4}}%
    \NewDocumentCommand{#1}{o o}%
    {%
        \IfNoValueTF{##1}%
        {%
            \mathOrText{#2}%
        }%
        {%
            \functionTemplate{#2}{\leftBracketType}{\rightBracketType}{##1}[##2]%
        }%
    }%
}

\DeclareDocumentCommand{\probabilisticFunctionTemplate}{m m O{} o}
{%
    \functionTemplate{#1}%
    {\lbrack}%
    {\rbrack}%
    {#2\IfEmptyTF{#3}{}{\ \IfNoValueTF{#4}{\left}{#4}\vert\ \vphantom{#2}#3\IfNoValueTF{#4}{\right.}{}}}%
    [#4]%
}

\ifboldNumberSets

    \newcommand*{\indicatorFunctionSymbol}{\mathbf{1}}
\else

    \newcommand*{\indicatorFunctionSymbol}{\mathds{1}}
\fi

\RenewDocumentCommand{\Pr}{m O{} o}%
{%
    \probabilisticFunctionTemplate{\mathrm{Pr}}{#1}[#2][#3]%
}

\NewDocumentCommand{\E}{m O{} o}%
{%
    \probabilisticFunctionTemplate{\mathrm{E}}{#1}[#2][#3]%
}

\NewDocumentCommand{\Var}{m O{} o}%
{%
    \probabilisticFunctionTemplate{\mathrm{Var}}{#1}[#2][#3]%
}

\DeclareDocumentCommand{\bigO}{m o}%
{%
    \functionTemplate{\mathrm{O}}{(}{)}{#1}[#2]%
}

\DeclareDocumentCommand{\smallO}{m o}%
{%
    \functionTemplate{\mathrm{o}}{(}{)}{#1}[#2]%
}

\DeclareDocumentCommand{\bigTheta}{m o}%
{%
    \functionTemplate{\upTheta}{(}{)}{#1}[#2]%
}

\DeclareDocumentCommand{\bigOmega}{m o}%
{%
    \functionTemplate{\upOmega}{(}{)}{#1}[#2]%
}

\DeclareDocumentCommand{\smallOmega}{m o}%
{%
    \functionTemplate{\upomega}{(}{)}{#1}[#2]%
}

\DeclareDocumentCommand{\eulerE}{o}%
{%
    \mathOrText{\mathrm{e}\IfNoValueTF{#1}{}{^{#1}}}%
}

\DeclareDocumentCommand{\poly}{m o}%
{%
    \functionTemplate{\mathrm{poly}}{(}{)}{#1}[#2]%
}

\createFunction{\id}{\mathrm{id}}

\NewDocumentCommand{\ind}{m o o}%
{%
    \IfNoValueTF{#2}%
    {%
        \mathOrText{\indicatorFunctionSymbol_{#1}}%
    }%
    {%
        \functionTemplate{\indicatorFunctionSymbol_{#1}}{(}{)}{#2}[#3]%
    }%
}

\DeclareDocumentCommand{\dom}{m o}%
{%
    \functionTemplate{\mathrm{dom}}{(}{)}{#1}[#2]%
}

\DeclareDocumentCommand{\rng}{m o}%
{%
    \functionTemplate{\mathrm{rng}}{(}{)}{#1}[#2]%
}

\DeclareDocumentCommand{\d}{o}%
{%
    \mathrm{d}\IfNoValueTF{#1}{}{^{#1}}%
}

\DeclareDocumentCommand{\set}{m m o}%
{
    \mathOrText{\IfNoValueTF{#3}{\left}{#3}\{#1\ \IfNoValueTF{#3}{\left}{#3}\vert\
    \vphantom{#1}#2\IfNoValueTF{#3}{\right.}{}\IfNoValueTF{#3}{\right}{#3}\}}
}      

\newcommand{\lofwidth}{0.64\linewidth} 

\begin{document}

    \frontmatter

\ifprintVersion
    \ifprofessionalPrint
        \newgeometry
        {
            textwidth = 134 mm,
            textheight = 220 mm,
            top = 38 mm + \extraborderlength,
            inner = 38 mm + \mybindingcorrection + \extraborderlength,
        }
    \else
        \newgeometry
        {
            textwidth = 134 mm,
            textheight = 220 mm,
            top = 38 mm,
            inner = 38 mm + \mybindingcorrection,
        }
    \fi
\else
    \newgeometry
    {
        textwidth = 134 mm,
        textheight = 220 mm,
        top = 38 mm,
        inner = 38 mm,
    }
\fi

\begin{titlepage}
    \sffamily
    \begin{center}
        \includegraphics[height = 3.2 cm]{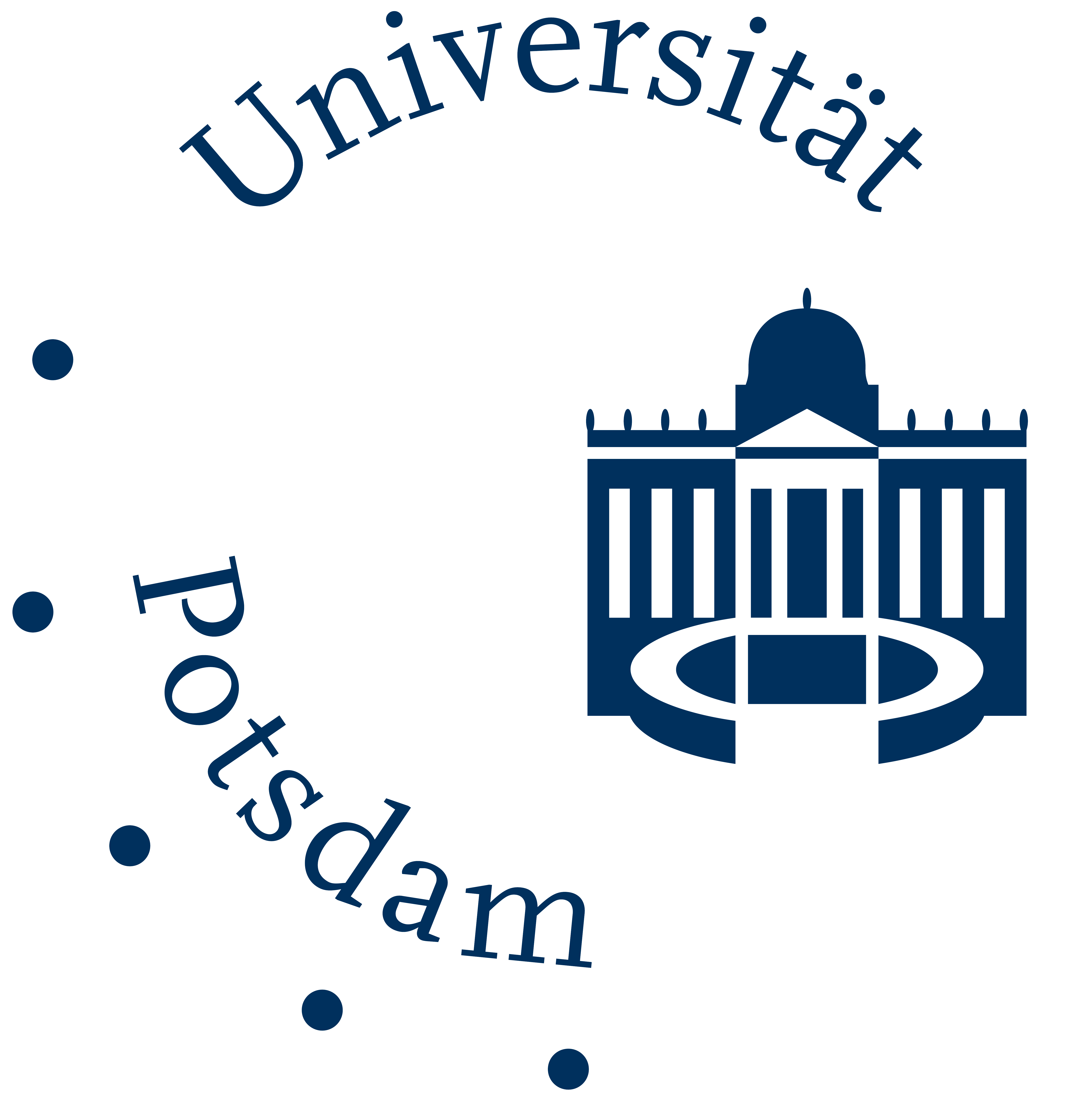} \hfill \includegraphics[height = 3 cm]{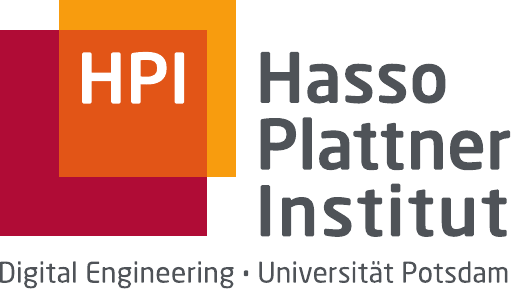}\\
        \vfil
        {\LARGE
            \rule[1 ex]{\textwidth}{1.5 pt}
            \onehalfspacing\printTitleBold\\[1 ex]
            {\vspace*{-1 ex}\Large \printGermanTitle}\\
            \rule[-1 ex]{\textwidth}{1.5 pt}
        }
        \vfil
        {\Large\textbf{\printAuthor}}
        \vfil
        {\large Universitäts\colloquialDegreeNameLowercase arbeit\\[0.25 ex]
        zur Erlangung des akademischen Grades}\\[0.25 ex]
        \bigskip
        {\Large \colloquialDegreeName{} of Science}\\[0.5 ex]
        {\large\emph{(\degreeAbbreviation\,Sc.)}}\\
        \bigskip
        {\large im Studiengang\\[0.25 ex]
        \printProgram}
        \vfil
        {\large eingereicht am \printDateReceived{} am\\[0.25 ex]
        Fachgebiet Algorithm Engineering der\\[0.25 ex]
        Digital-Engineering-Fakultät\\[0.25 ex]
        der Universität Potsdam}
    \end{center}
    
    \vfil
    \begin{table}[h]
        \centering
        \large
        \sffamily 
        {\def\arraystretch{1.2}
            \begin{tabular}{>{\bfseries}p{3.8 cm}p{5.3 cm}}
                Gutachter               & \printNameOfSupervisor\\
                Betreuer                & \printAdditionalExaminers
            \end{tabular}
        }
    \end{table}
\end{titlepage}

\restoregeometry

    \pagestyle{plain}

    \addchap{Abstract}

Outlier detection is a well-researched and crucial problem in machine learning. However, there is little research on string data outlier detection, as most literature focuses on outlier detection of numerical data. A robust string data outlier detection algorithm could assist with data cleaning or anomaly detection in system log files. In this thesis, we compare two string outlier detection algorithms. Firstly, we introduce a variant of the well-known \emph{local outlier factor} algorithm, which we tailor to detect outliers on string data using the \emph{Levenshtein} measure to calculate the density of the dataset. We present a differently weighted \emph{Levenshtein} measure, which considers hierarchical character classes and can be used to tune the algorithm to a specific string dataset. Secondly, we introduce a new kind of outlier detection algorithm based on the hierarchical left regular expression learner, which infers a regular expression for the expected data. Using various datasets and parameters, we experimentally show that both algorithms can conceptually find outliers in string data. We show that the regular expression-based algorithm is especially good at finding outliers if the expected values have a distinct structure that is sufficiently different from the structure of the outliers. In contrast, the local outlier factor algorithms are best at finding outliers if their edit distance to the expected data is sufficiently distinct from the edit distance between the expected data.

    \selectlanguage{ngerman}
    \addchap{Zusammenfassung}
    Ausreißererkennung ist ein gut erforschtes und sehr relevantes Problem im Bereich des maschinellen Lernens. Allerdings gibt es nicht viel Forschung zur Ausreißererkennung auf Zeichenketten, da sich die meiste Literatur auf numerische Ausreißererkennung fokussiert. Ausreißererkennung auf Zeichenketten kann beispielweise bei Datenbereinigung oder Anomalieerkennung in Systemprotokolldateien behilflich sein. In dieser Arbeit vergleichen wir zwei Ausreißererkennungsalgorithmen für Zeichenketten. Zunächst führen wir eine Variante des bekannten \emph{local outlier factor}-Algorithmus ein. Diese verwendet eine \emph{Levenshtein}-Metrik, um die Dichte des Datensatzes zu berechnen, sodass der Algorithmus Zeichenketten entgegennehmen kann. Außerdem präsentieren wir eine Option, die \emph{Levenshtein}-Metrik mit einer Hierarchie aus Zeichenklassen zu gewichten, um den Algorithmus besser an den konkreten Datensatz anpassen zu können. Der zweite Algorithmus ist eine neue Variante, Ausreißer auf Zeichenketten zu erkennen, die einen regulären Ausdruck für die erwarteten Daten erstellt. Mit verschiedenen Datensätzen und Parametern zeigen wir experimentell, dass beide Algorithmen konzeptuell Ausreißer in Zeichenkettendatensätzen erkennen können. Wir zeigen, dass der Algorithmus, der auf regulären Ausdrücken basiert, besonders gut Ausreißer erkennt, wenn die erwarteten Werte eine feste Struktur haben, die ausreichend von den Ausreißern unterschieden werden kann. Die \emph{local outlier factor}-Algorithmen sind hingegen besonders gut darin, Ausreißer zu erkennen, falls sich die Bearbeitungsdistanz zwischen Ausreißer und erwarteten Werten genug von der Bearbeitungsdistanz innerhalb der erwarteten Werte unterscheidet.
    \selectlanguage{american}

    \addchap{Acknowledgments}

First and foremost, I would like to thank my advisors, Timo and Xiaoyue, for our regular meetings, their constructive and quick feedback, and for helping me stay on track during the writing period.

A big thank you also goes out to Flora for the co-development of the quality report extraction tool and to Felix, with whom I shared the pleasure of working on outlier detection methods.

Thanks to my fellow bachelor's project students, Aleks, Antonia, Hendrik, Johanna, Moritz, and Paul, for being great office mates who made writing this thesis fun.

    \setuptoc{toc}{totoc}
    \tableofcontents

    \pagestyle{headings}
    \mainmatter

    \chapter{Introduction} \label{chapter:01_introduction}

Outlier detection is a fundamental problem in computer science that aims to identify anomalous data points in a dataset. Strategies to detect outliers date back to the 19\textsuperscript{th} century, with \cite{edgeworth_discordant_1887} being one of the first papers discussing outlier detection. Since then, there has been much more research on outlier detection, and researchers from many computer science disciplines have introduced various approaches to detect outliers. Survey papers such as \cite{hodge_survey_2004}, \cite{chandola_anomaly_2009}, or \cite{smiti_critical_2020} give an overview of these different outlier detection algorithms.

The definition of an outlier that should be detected varies from approach to approach. Many researchers build upon the outlier definition by the mathematician Hawkins \cite{hawkins_identification_1980}, who introduces them as ``[an] observation which deviates so much from other observations as to arouse suspicions that it was generated by a different mechanism''. An outlier is thereby characterized by being suspiciously and highly different from other available data or not fitting into the foundational concepts we have for the data (see also \cite{hodge_survey_2004}). 

The uses of outlier detection algorithms are manifold. Successful examples include the monitoring of fraudulent activity in credit card payments \cite{aleskerov_cardwatch_1997}, detecting landmines in ground satellite images \cite{byers_nearest-neighbor_1998}, or analysis of mammography images to detect some tumors \cite{spence_detection_2001}. The underlying algorithms stem from various computer science disciplines, including statistical approaches, machine learning-based approaches, nearest neighbor-based approaches, clustering-based approaches, or information-theoretic approaches. While the research on outlier detection and the approaches are extensive, the focus lies primarily on numerical data, with most algorithms only accepting numbers as inputs (see \cite{hodge_survey_2004}, \cite{chandola_anomaly_2009}). Nevertheless, detecting outliers on string data is also a highly relevant topic, as it makes ground for automated data cleaning on user input data, analysis of protein sequences, or detecting anomalous activity in system log files.

Our goal with this thesis is to find outlier detection algorithms that reliably detect outliers in string data. To achieve that goal, we present two different outlier detection algorithms.
The first algorithm builds upon the commonly used $k$-nearest neighbor algorithm called \emph{local outlier factor}, as presented in \cite{breunig_lof_2000}. It uses distances between data points to measure each data point's similarity. The general idea is that data outside dense areas is probably anomalous, while dense areas probably contain expected data. This approach finds density-based outliers and can be adapted to accept string data using a string distance metric such as the Levenshtein metric.
The second algorithm is a new concept based on the regular expression learner presented in \cite{doskoc_efficient_2016}. The idea relies on the assumption that we can describe the expected data with a specific language represented by a regular expression. Therefore, we try to find a regular expression that defines the expected data, and label every data point not in its language as an outlier. 
Using these two algorithms, we explore whether they can reliably detect outliers in string data. Building on that, we analyze what kind of string dataset works best with which algorithm, and how those two algorithms compare. To answer these questions, we use synthetic and real-world string datasets. We use standard evaluation metrics, such as the receiver operating characteristic, to evaluate and compare their performance. We hope both algorithms can successfully detect outliers in string data, but assume they differ in the dataset on which they perform best.

The thesis is structured as follows. In \Cref{chapter:02_string_data}, we introduce the type of string data in which we want to detect outliers and what kind of outliers we want to detect. \Cref{chapter:03_k_nn_levenshtein} formally introduces the \emph{local outlier factor} algorithm we modified to detect outliers on string data. We introduce the second regular expression-based algorithm in \Cref{chapter:04_regex_based_approach}. In \Cref{chapter:05_using_real_world_data}, we compare those two algorithms on various real-world data sets, highlighting their similarities and differences. We conclude our findings in \Cref{chapter:06_conclusions}.

\section{Related Work}
Outlier detection algorithms are a well-researched topic, and similar to this thesis, many papers compare different outlier algorithms on various datasets.

Researchers in \cite{domingues_comparative_2018} use different publicly available numerical industry datasets to compare various unsupervised machine learning algorithms for outlier detection. They compare different algorithms using receiver operating characteristic charts and precision-recall charts, but restrict themselves to numerical data.

Similar research gets presented in \cite{campos_evaluation_2016}, focusing on $k$-nearest neighbor algorithms. In this paper, researchers extensively discuss numerical datasets and their suitability for benchmarking outlier detection algorithms, and evaluate different measures that aim to assess the performance of the algorithms.

A focus on comparing statistical outlier detection methods provides \cite{bakar_comparative_2006}, where researchers compare control chart techniques with linear regression and a $k$-nearest neighbor algorithm that uses the Manhattan distance to detect outliers in a large air pollution dataset.

In \cite{xu_comparison_2018}, researchers compare outlier detection algorithms focusing on detecting outliers in high-dimensional numerical data. They also use $k$-nearest neighbor-based approaches in addition to subspace and ensemble learning-based methods, and provide values for the area under the receiver operating characteristic curve, the precision of the algorithms, and their rank power.

    \chapter{Specification of String Data for Outlier Detection} \label{chapter:02_string_data}
    To properly compare and evaluate outlier detection algorithms, we first need to define the scope and nature of the strings on which we want to detect outliers. This chapter elaborates on that, using the intuitive outlier definition by \cite{hawkins_identification_1980} as described in the introduction.

We have two options regarding what output the outlier detection algorithm should produce. The first option is categorical outlier detection, where a data point gets labeled as either an outlier or an expected value, without any confidence value or score that denotes the degree to which the datum is an outlier. That differs from the second option, gradual outlier detection, which assigns an anomaly score to every data point. The higher the score, the more anomalous the data point is and therefore the more likely it is that this data point is an outlier \cite{hodge_survey_2004}. A categorical outlier detection algorithm often uses a gradual method and adds a threshold function to classify anomaly scores as outliers or non-outliers. We also use this approach in this thesis, applying a thresholding function to the anomaly scores we receive from the algorithms to achieve categorical outlier detection. We need categorical outlier detection to compare the algorithms in \Cref{chapter:05_using_real_world_data}, as it makes evaluating the correctness and accuracy of the outlier detection algorithms easier.

Furthermore, we must define the kind of string data we are considering. Generally, a string data point consists of zero, one, or multiple consecutive characters $c\in\Sigma$, where $\Sigma$ is the alphabet of the possible characters and $\epsilon$ represents the empty string.
We have two options to declare the information complexity of the string data. First, we could declare that every string contains exactly one word. With this assumption, we would treat it as a single enclosed datum without differentiating between different string parts. On the other hand, we could assume that the string may contain multiple words. In that case, there exists a separation character $\lambda\in\Sigma$ that separates the various parts of the string. A word is a sequence of characters $c\in\Sigma\setminus\{\lambda\}$, and a datum comprises one or multiple words. Here, the outlier detection algorithm must account for the different words, which would require different approaches than detecting outliers on single-word strings. Strings of this category are typically longer than single-word strings. Examples include sentences, paragraphs, or newspaper articles. For this thesis, we will focus on strings that contain exactly one word and utilize outlier detection algorithms specifically designed for this use case.

We could also have two different approaches concerning the amount of information we consider when detecting outliers. We can perform outlier detection either solely based on the syntactical properties of the data, or we could integrate a context in addition to the data itself to evaluate its semantics.
The first syntax-only approach to outlier detection does not require additional data to detect outliers. However, one may choose a more suitable algorithm for a specific problem. An example is a set of strings that consists mostly of ISO 8601 date strings (see \cite{deutsches_institut_fur_normung_din_2020} for a definition) and some dates formatted differently, serving as outliers in our set. Let $D_1=\{\text{``2025-09-30'', ``2025-04-01'', ``22nd of April 2004''}\}$ be an excerpt of the data. A good syntactical outlier detection algorithm should detect differently formatted strings as outliers, in this case, ``22nd of April 2004''. It can detect this outlier simply by evaluating the string data itself. It does not need the information that these strings represent a date and thus does not need additional context to detect outliers. In contrast, syntactical outlier detection requires additional context to classify outliers correctly. Take, for example, the string set $D_2=\{$``red'', ``green'', ``blue'', ``glue''$\}$. Here, most strings describe a color, except ``glue'' which the algorithm should consider an outlier. However, the algorithm needs additional context to detect this outlier. In this case, at least information about which string represents a color, which, for example, could be determined using a word cloud. A pure syntactical approach would not suffice, as the information about which strings belong together is not given by its syntax, but rather by interpretation with the additional context.
One could also think about hybrid gradual approaches, which weigh their anomaly score by considering both the anomaly score in syntax and semantics. For this thesis, we will focus on syntactical outlier detection without requiring additional context to identify outliers.

To conclude, there are different possibilities for detecting outliers in string data. One could consider categorical labeling or gradual scoring; the strings may contain one or multiple words, and the outlier detection algorithm might attempt to identify syntactical or semantic outliers. This thesis focuses on syntactical outlier detection for string data, where each data point contains exactly one word.

    \chapter[K-Nearest Neighbor-Based Approach]{K-Nearest Neighbor-Based \\Approach}
    \label{chapter:03_k_nn_levenshtein}
    Our first approach to outlier detection on string data utilizes a traditional $k$-nearest neighbor algorithm. The $k$-nearest neighbor algorithms are a family of outlier detection algorithms that use a distance measure between the data points to infer an anomaly score for each datum, which the algorithm calculates relative to the $k$-nearest data point. The higher the anomaly score, the more the algorithm assesses the datum as an outlier. 

Researchers in \cite{ramaswamy_efficient_2000} introduce one of the first and most commonly cited $k$-nearest neighbor outlier algorithms. It uses a datum's distance to its $k^{th}$ nearest neighbor as the anomaly score, thus identifying \textit{distance}-based outliers. Another kind of outliers to detect with $k$-nearest neighbor algorithms are \textit{density}-based outliers. One commonly cited algorithm to identify \textit{distance}-based outliers is the local outlier factor (LOF) algorithm introduced in \cite{breunig_lof_2000}. It calculates a density relative to the $k$ nearest other data points for every datum in the dataset. The algorithm then calculates a local outlier factor (LOF) for every datum, which it uses as the anomaly score. The LOF is equal to the average of the local densities of other data of the $k$ nearest data points divided by the density of the current datum. Essentially, it encompasses the average ratio of the local reachability density of a datum to the local reachability densities of its $k$-nearest neighbors. We formally introduce the algorithm in \Cref{sec:03_01}.

Whether one should use the \textit{density}-based algorithm or the \textit{distance}-based algorithm depends on the concrete problem at hand and the nature of the dataset. In practice, the \textit{density}-based outlier detection algorithm outperforms the \text{distance}-based outlier detection algorithm in various sample datasets, as shown in \cite{yang_outlier_2023}. Therefore, we use the LOF-variant as our $k^{th}$-nearest neighbor algorithm of choice.

\section{Formal Introduction of the Local Outlier Factor (LOF) Algorithm}
\label{sec:03_01}
This section introduces relevant definitions and formulas for calculating the Local Outlier Factor (LOF), as described in \cite{breunig_lof_2000}. 

Let $D$ be a set of strings, $k\in\mathbb N^+$ with $1\le k<|D|$ the number of neighbors to consider, and $d(D\times D)\to \mathbb N$ a distance measure. First, we define the $k$-distance neighborhood of an object $p\in D$, followed by the reachability distance of an object $p\in D$ to an object $o\in D$. Based on these definitions, we define the local reachability density $\operatorname{lrd}$ of an object $p\in D$, and, with that, the local outlier factor $\operatorname{lof}$ of an object $p\in D$, which will be our anomaly score.

\begin{definition}[$k$-distance Neighborhood of a String $p$]
    \label{def:03_distance_neighborhood}
    Let $p\in D$ be a string from the dataset. We define the $\text{$k$-distance}$ as follows.
    \[\text{$k$-distance}(p)=\inf_{r \in \mathbb{R_+}} \; \left|\{o \in D \mid d(p,o) \leq r\}\right| \leq k\]
    Note that at least $k$ data points have a distance equal to or less than $\text{$k$-distance}(p)$ to $p$, and that there are at most $k-1$ data points, which have a distance less than $\text{$k$-distance}(p)$ to $p$.
    
    Given this $k$-distance definition, we define the $k$-distance neighborhood $N_k(p)$ of a string $p$ as follows.
    \[ N_k(p)=\{q\in D\setminus \{p\}~|~d(p,q)\le k\text{-distance}(p)\}\].     
    Based on the note above, the cardinality of $N_k(p)$ must not strictly be $k$. It might be larger than $k$ if multiple objects have the same $\text{$k$-distance}(p)$ to $p$. From now on, we call $N_k(p)$ the $k$-nearest neighbors of $p$.
\end{definition}

Building on the $k$-distance neighborhood of a string $p\in D$, we define the reachability distance of a string $p$ to another string $o$. It represents the furthest distance between $p$ and $o$, as seen below.

\begin{definition}[Reachability Distance of an Object $p$ With Respect to an Object $o$]
    Let $p, o\in D, p\ne o$ be two different strings from the dataset. We define the \textit{reachability distance} of $p$ to $o$ as follows.
    \[\operatorname{reach-dist}_k(p,o)=\max\{k\text{-distance}(o),d(p,o)\}\]
    We recommend Figure 2 in \cite{breunig_lof_2000} for an explanatory illustration.
\end{definition}

With these definitions, we can calculate the local reachability density of a string $p\in D$. It represents the density of all strings in the reachability distance to $p$. 

\begin{definition}[local reachability density of a string $p$]
    Let $p\in D$ be a string from the dataset. The local reachability density is the inverse of the average reachability distance based on the $k$-nearest neighbors of $p$.
    \[\operatorname{lrd}_{k}(p)=\frac{\left|N_{k}(p)\right|}{\sum_{o\in N_{k}(p)}\operatorname{reach-dist}_{k}(p,o)}\]
\end{definition}

To ensure that $\operatorname{lrd}$ is valid and we do not divide by zero, the sum of all reachability distances must not be zero. To circumvent that, we assume that the set of data points $D$ only contains unique objects as proposed in \cite{breunig_lof_2000}. A dataset with duplicate items must thus be made unique before running the algorithm.

At last, we define the Local Outlier Factor (LOF) of a string $p\in D$, which this algorithm uses as the anomaly score.

\begin{definition}[Local Outlier Factor (LOF) of a String $p$]
    \label{def:03_lof}
    The \textit{(local) outlier factor} of $p$ is defined as
    \[\operatorname{LOF}_{k}(p)=\frac{\sum_{o\in N_{k}(p)}{\frac{\operatorname{lrd}_{k}(o)}{\operatorname{lrd}_{k}(p)}}}{\left|N_{k}(p)\right|}\]
\end{definition}

\section{Distance Measures for Strings}
\label{sec:03_distance_measures_for_strings}
Like all $k$ nearest neighbor algorithms, the LOF algorithm needs a measure to calculate the distance between two data points. One suitable distance measure for string data is an edit distance, specifically the Levenshtein distance, as explained in \cite{navarro_guided_2001}. In its original version, the costs of inserting, replacing, and deleting a character in a word are equally weighted with $+1$. But equal weights are not always ideal when comparing string data, as the following example shows.

\begin{example}[Levenshtein Measure With Default Weights]
    Let $D_3=\{\text{``$2000$-$01$-$01$''}, \text{``$1999$-$12$-$30$''}, \text{``$200$th time''}\}$. We assume that $D_3$ should only contain strings in the ISO 8601 date format (see \cite{deutsches_institut_fur_normung_din_2020}) and that ``$200$th time'' is an outlier. As $k$-nearest neighbor algorithms rely on the distances between data points to calculate the anomaly score, we would like the distance between the ISO 8601 date strings to be small, and the distance from those to our outlier to be high. Looking at the basic Levenshtein distances between those strings, we get $8$ as a distance for the two ISO 8601 date strings (see \Cref{tab:03_d_between_two_iso_date_strings}), but only $7$ between one of the ISO 8601 dates and our outlier (see \Cref{tab:03_d_between_a_iso_date_string_and_the_outlier}). As we hoped, the other pairing has a relatively high distance of $10$ (see \Cref{tab:03_d_between_the_other_iso_date_string_and_the_outlier}). To foster a better understanding, we note the weighting operation of each character above its weight in the tables, with \texttt{R} representing the replacement weight and $=$ indicating that the character remains unchanged.

    \begin{table}[ht!]
        \centering
        \begin{tabular}{|ccccccccccc|}
            \hline
             2 & 0 & 0 & 0 & - & 0 & 1 & - & 0 & 1 & \\
             1 & 9 & 9 & 9 & - & 1 & 2 & - & 3 & 0 & \\
             \hline
             \texttt{R} & \texttt{R} & \texttt{R} & \texttt{R} & \texttt{=} & \texttt{R} & \texttt{R} & \texttt{=} & \texttt{R} & \texttt{R} & \\
             \hline
             +1 & +1 & +1 & +1& 0 & +1 & +1 & 0 & +1 & +1 & = 8 \\
            \hline
        \end{tabular}
        \caption{Distance between the two ISO 8601 date strings}
        \label{tab:03_d_between_two_iso_date_strings}
    \end{table}
    
    \begin{table}[ht!]
        \centering
        \begin{tabular}{|ccccccccccc|}
             \hline
             2 & 0 & 0 & 0 & - & 0 & 1 & - & 0 & 1 & \\
             2 & 0 & 0 & t & h &   & t & i & m & e &  \\
             \hline
             \texttt{=} & \texttt{=} & \texttt{=} & \texttt{R} & \texttt{R} & \texttt{R} & \texttt{R} & \texttt{R} & \texttt{R} & \texttt{R} & \\
             \hline
             0 & 0 & 0 & +1 & +1 & +1 & +1 & +1 & +1 & +1 & = 7 \\
             \hline
        \end{tabular}
        \caption{Distance between an ISO 8601 date string and the outlier}
        \label{tab:03_d_between_a_iso_date_string_and_the_outlier}
    \end{table}
    
    \begin{table}[ht!]
        \centering
        \begin{tabular}{|ccccccccccc|}
             \hline
             1 & 9 & 9 & 9 & - & 1 & 2 & - & 3 & 0 & \\
             2 & 0 & 0 & t & h &   & t & i & m & e & \\
             \hline
             \texttt{R} & \texttt{R} & \texttt{R} & \texttt{R} & \texttt{R} & \texttt{R} & \texttt{R} & \texttt{R} & \texttt{R} & \texttt{R} & \\
             \hline
             +1 & +1 & +1 & +1 & +1 & +1 & +1 & +1 & +1 & +1 & = 10 \\
             \hline
        \end{tabular}
        \caption{Distance between the other ISO 8601 date string and the outlier}
        \label{tab:03_d_between_the_other_iso_date_string_and_the_outlier}
    \end{table}

    The default costs for the Levenshtein distance yield quite similar distance measures for the three strings, with which we cannot reliably detect ``$200$th time'' as an outlier. Adjusting the weights for the three basic operations \textit{inserting}, \textit{replacing}, and \textit{deleting} does not solve our problem. If we change the weights, the distribution of the distance values remains, as we only use the weights of the replacing operation. It would only be a factor for the result and thus not helpful, as the distance between our normal values falls between the distances to the outlier.
\end{example}

To correctly identify the outlier in such cases, we need a more distinct weighting approach for the distances, particularly for the replacement operation. As a basis, we use the following observation. When analyzing the syntactical properties of a string, we can map some characters into specific classes. We could, for example, have one class for digits, one for punctuation, one for lowercase alphabetic characters, and one for uppercase alphabetic characters. These classes can then form higher-level character classes themselves. For example, the lowercase alphabetic characters and the uppercase alphabetic characters belong to the class of alphabetic characters, and the numbers and alphabetic characters belong to the class of alphanumeric characters. We can therefore create a hierarchical partition tree of character classes as defined in \cite{doskoc_efficient_2016}.

\label{def:03_hierarchical_partitioning}
\begin{definition}[Hierarchical Partition]
    Let $\Sigma$ be an alphabet. Then $P\subset \mathcal{P}(\Sigma)$ is called a \textbf{hierarchical partition} of $\Sigma$ if it fulfills the following properties.
    \begin{enumerate}[(a)]
        \item $\Sigma \in P$.
        \item $\forall c\in \Sigma: \{c\} \in P$.
        \item $\forall p,q\in P: p\cap q\in\{p,q,\emptyset\}$.
    \end{enumerate}
\end{definition}

\begin{figure}
    \centering
    \begin{forest}
    for tree={
      draw,
      rounded corners,
      align=center,
      edge={->},
      parent anchor=south,
      child anchor=north,
      l sep=15pt,
      s sep=10pt,
    }
    [$\Sigma$
        [a-zA-Z0-9\textvisiblespace.{,}-
            [a–zA–Z0–9
                [a–zA–Z
                    [a–z
                        [a]
                        [b]
                        [\dots]
                        [z]
                    ]
                    [A–Z
                        [A]
                        [B]
                        [\dots]
                        [Z]
                    ]
                ]
                [0–9
                    [0]
                    [1]
                    [\dots]
                    [9]
                ]
            ]
            [\textvisiblespace.{,}-
                [\textvisiblespace]
                [.]
                [{,}]
                [-]
            ]
        ]
        [@]
        [+]
        ["]
        [\dots]
    ]
    \end{forest}
    \caption{Example hierarchy $H$ tailored to date outlier detection for use in a weighted Levenshtein metric}
    \label{fig:03_proposed_hierarchy}
\end{figure}
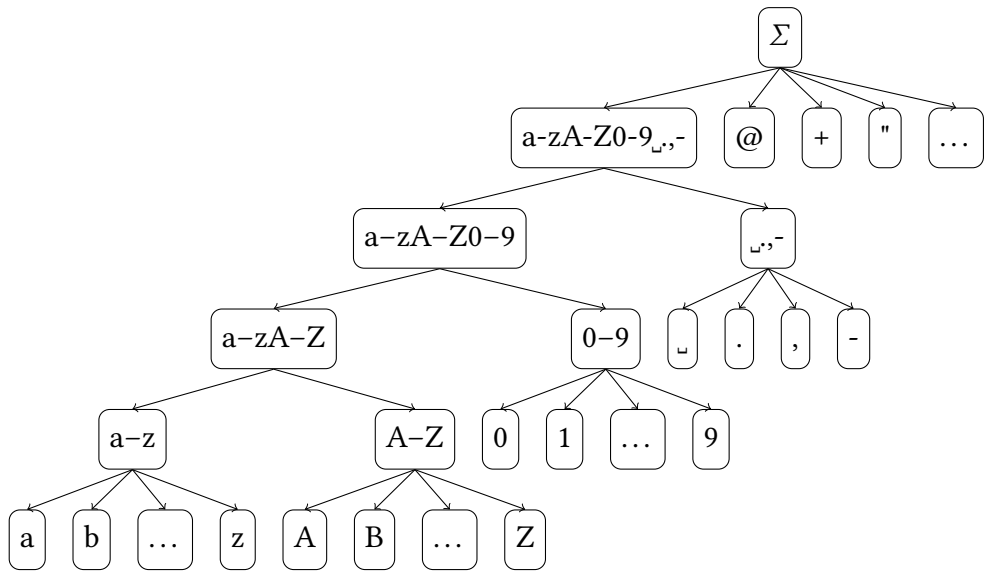

To tailor this definition to our use case of outlier detection, we will define the alphabet $\Sigma$ as the set that contains every character of every string in the dataset we examine. In the hierarchy definition, we denote the remaining unlisted characters from the alphabet $\Sigma$ with $\dots$, as shown in \Cref{fig:03_proposed_hierarchy}. We can then use a hierarchical partition of this kind as an input for the weighting function to better reflect the syntactical properties of our strings in the distance measure. The syntactical properties of the data depend on the use case and nature of the dataset, so the hierarchy needs to reflect the expectations one has of the dataset itself. Hence, it can be a crucial parameter for tailoring the outlier detection algorithm to a specific dataset, thereby improving the accuracy of outlier labeling.

As an example for such a hierarchy, we propose the hierarchy $H$ in \Cref{fig:03_proposed_hierarchy}, which we have tailored to outlier detection for date strings. It has character classes for the lowercase alphabetic characters, the uppercase alphabetic characters, the complete alphabetic characters, the numerical characters, the alphanumerical characters, and all other possible characters in the alphabet $\Sigma$ of our string data $D$. Possible date formatting characters, such as ``\textvisiblespace'', ``.'', ``,'' and ``-'' are in one class and more closely connected to the alphanumeric characters. All other characters are direct children of the root node. Instead of using $1$ as the weight for the replacing operation in the Levenshtein metric, we now use the path length between the two characters in the hierarchy tree $H$ as the weight for the operation. This way, the syntactical properties of our string data are considered when calculating the distance between data points, as this measure benefits replacement operations that stay within the same character class and punishes replacements of characters with far-apart character classes.

\begin{example}[Levenshtein Measure With Hierarchical Weights]
    We use $D_3$ from the previous example, but now weight the operations based on the hierarchy $H$. With this measure, the two ISO 8601 strings are closer together (see \Cref{tab:03_h_d_between_two_iso_date_strings}), while both ISO 8601 strings now have a larger distance to the outlier string (see \Cref{tab:03_h_d_between_a_iso_date_string_and_the_outlier} and \Cref{tab:03_h_d_between_the_other_iso_date_string_and_the_outlier}).
\end{example}

\begin{table}
    \centering
    \begin{tabular}{|ccccccccccc|}
         \hline
         2 & 0 & 0 & 0 & - & 0 & 1 & - & 0 & 1 & \\
         1 & 9 & 9 & 9 & - & 1 & 2 & - & 3 & 0 & \\
         \hline
         \texttt{R} & \texttt{R} & \texttt{R} & \texttt{R} & \texttt{=} & \texttt{R} & \texttt{R} & \texttt{=} & \texttt{R} & \texttt{R} & \\
         +0.5 & +0.5 & +0.5 & +0.5 & 0 & +0.5 & +0.5 & 0 & +0.5 & +0.5 & = 4 \\
         \hline
    \end{tabular}
    \caption{Hierarchical weighted distance between the two ISO 8601 date strings}
    \label{tab:03_h_d_between_two_iso_date_strings}
\end{table}

\begin{table}
    \centering
    \begin{tabular}{|ccccccccccc|}
         \hline
         2 & 0 & 0 & 0 & - & 0 & 1 & - & 0 & 1 & \\
         2 & 0 & 0 & t & h &   & t & i & m & e &  \\
         \hline
         \texttt{=} & \texttt{=} & \texttt{=} & \texttt{R} & \texttt{R} & \texttt{R} & \texttt{R} & \texttt{R} & \texttt{R} & \texttt{R} & \\
         0 & 0 & 0 & +1.25 & +1.5 & +1.25 & +1.25 & +1.5 & +1.25 & +1.25 & = 9.25 \\
         \hline
    \end{tabular}
    \caption{Hierarchical weighted distance between an ISO 8601 date string and the outlier}
    \label{tab:03_h_d_between_a_iso_date_string_and_the_outlier}
\end{table}

\begin{table}
    \centering
    \begin{tabular}{|ccccccccccc|}
         \hline
         1 & 9 & 9 & 9 & - & 1 & 2 & - & 3 & 0 & \\
         2 & 0 & 0 & t & h &   & t & i & m & e & \\
         \hline
         \texttt{R} & \texttt{R} & \texttt{R} & \texttt{R} & \texttt{R} & \texttt{R} & \texttt{R} & \texttt{R} & \texttt{R} & \texttt{R} & \\
         +0.5 & +0.5 & +0.5 & +1.25 & +1.5 & +1.25 & +1.25 & +1.5 & +1.25 & +1.25 & = 10.75 \\
         \hline
    \end{tabular}
    \caption{Hierarchical weighted distance between the other ISO 8601 date string and the outlier}
    \label{tab:03_h_d_between_the_other_iso_date_string_and_the_outlier}
\end{table}

\section{Choosing a Value for k}
Before we can run the LOF algorithm, we need to set a value for the neighborhood size $k$. Researchers employ several approaches to select $k$ throughout the literature, as described in \cite{yang_outlier_2023}. Often, $k$ is chosen dependent on the input size $n$, for example $k=5\cdot log(n)$ in \cite{yuhua_li_selecting_2011}, $k=0.03 \cdot n$ in \cite{yang_mean-shift_2021}, or just set to a constant value, for example $k=30$ in \cite{pei_efficient_2006}. While these approaches may work well on the experimental data in those papers, they lack justification according to the researchers in \cite{yang_outlier_2023}. Instead, they propose an algorithm called $KFCS$ ($k$ finder based on neighborhood consistency, scoring-variant) to guess a value for $k$.

Their algorithm follows the assumption that an object $p$ should generally have a similar outlier score as its immediate $k$ neighborhood. They calculate a scoring-based neighborhood consistency value, defined below, to quantify this.

\begin{definition}[Scoring-Based Neighborhood Consistency]
    Let $D$ be a unique list of strings and $k\in\mathbb N$ with $1\le k< |D|$. We use the $k$-distance neighborhood $N_k(p)$ of an element $p\in D$ as defined in \Cref{def:03_distance_neighborhood} and the definition of the local outlier factor (LOF) $\operatorname{LOF}_k(p)$ for every $p\in D$ following \Cref{def:03_lof}. We define a list of scores $U_k$ as follows.
    \[U_k=\left\{\frac 1k\sum_{o\in N_k(p)}\operatorname{LOF}_k(o) \mid  p\in D\right\}\]
    Let $V_k$ be the list of all $\operatorname{LOF}_k(p)$ for every $p\in D$. The scoring-based neighborhood consistency $c_k$ is the inverse of the cosine similarity of $U_k$ and $V_k$, defined as follows.
    \[c_k=1-\cos(U_k,V_k)\]
\end{definition}

To complete the $KFCS$ guesser, we select the $k$ out of every possible $k\in \mathbb N$ with $1\le k<n$ which yields the highest \textit{scoring-based neighborhood consistency} value.

\begin{definition}[KFCS Guesser]
    Let $K=\{1\le n<|D|\mid n\in\mathbb N\}$ be the set of all possible $k$ values for a dataset $D$, and $c_k$ the \textit{scoring-based neighborhood consistency} value for a $k\in K$. The guess for a $k$ from the $KFCS$ guesser is defined as follows.
    \[k_{KFCS}={\arg \max}_{k\in K}c_k\]
\end{definition}

We will use this $KFCS$ guesser for $k$, as it performs reasonably well in the researchers' experimental tests compared to other approaches. For a complete description of the $KFCS$ algorithm and the other ranking-based variant see \cite{yang_outlier_2023},

\section{Setting a Threshold for the Anomaly Score}
Having a distance measure and a guesser for the value of $k$, we can now calculate the anomaly values for every data point. However, to classify the data into outliers and non-outliers based on their anomaly score, we also need a threshold value separating anomalous values from the normal ones. A commonly used approach is to define the threshold as a constant value. However, we use a different, more dynamic way of calculating the threshold to accommodate multiple outlier groups.

First, we calculate the mean $m$ of the anomaly values and set the threshold value $t$ to a multiple $f$ of that mean, with $t=f\cdot m$. We could, for example, use a factor of $f=2$, meaning that the threshold value will be twice the mean anomaly score, so $t=2\cdot m$. The algorithm now labels all data points with an anomaly value higher than $t$ as outliers, and removes them from the dataset. We then repeat this algorithm, which thus calculates a new mean and therefore a new threshold. This strategy incrementally detects new outliers and continues as long as the algorithm can classify new data points as outliers. Generally, the higher the threshold factor, the more data points the algorithm detects as outliers. This strategy thus accommodates multiple outlier groups with varying degrees of anomalousness.

In our experiments, we use this approach to thresholding with different values for $t$ and explore how they change the amount and the kind of outliers the algorithm detects.

\section{Initial Synthetic Experiments}
We use two synthetic datasets for an initial test of the LOF algorithm and its configurations. First, a clean dataset on which we do not expect any outliers, and a dirty one on which we expect the algorithm to find outliers. Both datasets have a total size of one thousand elements, with the second consisting of one percent of outlier values of varying degrees. The clean dataset consists of one thousand consecutive date strings, starting at ``2020-01-01''. The dirty dataset is a variant of the clean dataset, where we replaced ten values with outliers of various degrees from \Cref{tab:03_new_values_in_the_data_set}. Note that neither dataset contains duplicates, so we do not need to make them unique before running the LOF algorithm.

\begin{table}[ht]
    \centering
    \begin{tabular}{|r|p{95mm}|}
        \hline
        \textbf{Value} & \textbf{Description} \\
        \hline
        This is an outlier & syntactically different \\
        \hline
        \textit{empty string} & empty string, does not fit the ISO 8601 pattern \\
        \hline
        22nd of April 2004 & date in natural language \\
        \hline
        30.09.2025 & date formatted DD.MM.YYYY \\
        \hline
        01042024 & date without the hyphens \\
        \hline
        2012/01/01 & date formatted YYYY/MM/DD \\
        \hline
        0000-00-00 & not a valid date yet correctly formatted \\
        \hline
        2099-99-99 & also not a valid date, but correctly formatted and closer to the other dates \\
        \hline
        2000-01-01 & valid date but not in range \\
        \hline
        1999-12-31 & another valid date, but also not in range \\
        \hline
    \end{tabular}
    \caption{Values we have introduced into the dataset to test the outlier detection algorithm, sorted decreasingly by expectancy to be considered an outlier}
    \label{tab:03_new_values_in_the_data_set}
\end{table}

The experiment includes two configurations for the LOF algorithm, one with the equally weighted Levenshtein distance measure and one with the heuristically weighted Levenshtein distance measure, as described in the section above. Both configurations use the $KFCS$ guesser to guess a value for $k$, and the dynamic thresholding algorithm with $1.5, 2, 3$, and $5$ as threshold factors.

The first observation from the simulation is based on the results from the unmodified dataset, as shown in \Cref{tab:03_results_outlier_detection_unmodified}. Neither algorithm finds any outlier with the threshold factors, except for the smallest factor $1.5$, where there are some false positives. There is no real difference in the behavior of the two algorithm variants, except in the order of false positives the algorithms detect with a threshold factor of $1.5$. This result is to be expected, as all data in the clean dataset follows the same general structure, so there is no difference in using the default metric compared to the heuristically weighted variant.

The results from the modified dataset are more interesting, as shown in \Cref{tab:03_results_outlier_detection_modified}. Notably, neither algorithm labels an unmodified string as an outlier, resulting in zero false positives. Altogether, the number of data points labeled as outliers and those labeled as outliers varies between the weighted and unweighted Levenshtein distance measures and the different threshold factors. Generally, the algorithm with the unweighted Levenshtein distance finds outliers even at higher threshold factors. Both algorithms find all outliers from \Cref{tab:03_new_values_in_the_data_set} with the smallest threshold factor.

Like with the clean dataset, the structure of most data points is similar, following the ISO 8601 standard. Therefore, we only observe a difference in the order of the detected outliers. With later experiments, the difference between the two metrics is more prevalent, see \Cref{chapter:05_using_real_world_data}.

In conclusion, the density-based local outlier factor combined with the $KFCS$ $k$-value guesser and the multiplicative threshold approach can successfully detect most outliers in our sample string data set. The results differ between using a heuristically weighted Levenshtein distance measure and the standard, non-weighted distance measure, and the threshold factor used. One can tailor the distance measure and the threshold factor to suit the specific use case and dataset, improving the outlier detection algorithm results. In \Cref{chapter:05_using_real_world_data}, we will test this algorithm with various real-world data sets and compare it to the algorithm presented in \Cref{chapter:04_regex_based_approach}.

\begin{table}[hb]
    \centering

    \begin{tabular}{|c|p{50mm}|p{50mm}|}
        \hline
        \textbf{Threshold Factor} & \textbf{Unweighted Levenshtein} & \textbf{Weighted Levenshtein} \\
        \hline
        ``5'' & \textit{no outliers} & \textit{no outliers} \\
        \hline
        ``3'' & \textit{no outliers} & \textit{no outliers} \\
        \hline
        ``2'' & \textit{no outliers} & \textit{no outliers} \\
        \hline
        ``1.5'' & ~~\llap{\textbullet}~~``2020-04-01'' \newline ~~\llap{\textbullet}~~``2020-06-01'' \newline ~~\llap{\textbullet}~~``2022-09-08'' \newline ~~\llap{\textbullet}~~``2020-09-01'' \newline ~~\llap{\textbullet}~~``2022-02-09'' \newline ~~\llap{\textbullet}~~``2021-09-29'' \newline ~~\llap{\textbullet}~~``2022-09-09'' \newline ~~\llap{\textbullet}~~``2022-09-22'' \newline ~~\llap{\textbullet}~~``2022-09-23'' \newline ~~\llap{\textbullet}~~``2022-09-24'' \newline ~~\llap{\textbullet}~~``2022-09-25'' \newline ~~\llap{\textbullet}~~``2022-09-26'' \newline ~~\llap{\textbullet}~~``2022-09-27'' \newline ~~\llap{\textbullet}~~``2022-04-29'' \newline ~~\llap{\textbullet}~~``2022-06-29'' \newline ~~\llap{\textbullet}~~\dots & ~~\llap{\textbullet}~~``2020-05-31'' \newline ~~\llap{\textbullet}~~``2020-03-31'' \newline ~~\llap{\textbullet}~~``2020-07-31'' \newline ~~\llap{\textbullet}~~``2020-08-31'' \newline ~~\llap{\textbullet}~~``2021-08-31'' \newline ~~\llap{\textbullet}~~``2021-05-31'' \newline ~~\llap{\textbullet}~~``2021-07-31'' \newline ~~\llap{\textbullet}~~``2021-03-31'' \newline ~~\llap{\textbullet}~~``2021-06-11'' \newline ~~\llap{\textbullet}~~``2021-04-11'' \newline ~~\llap{\textbullet}~~``2020-04-11'' \newline ~~\llap{\textbullet}~~``2020-06-11'' \newline ~~\llap{\textbullet}~~``2020-09-30'' \newline ~~\llap{\textbullet}~~``2021-09-30'' \newline ~~\llap{\textbullet}~~``2020-06-22'' \newline~~\llap{\textbullet}~~\dots
        \\
        \hline
    \end{tabular}
    
    \caption{Detected outliers in the unmodified dataset for different thresholds}
    \label{tab:03_results_outlier_detection_unmodified}
\end{table}

\begin{table}
    \centering

    \begin{tabular}{|c|p{50mm}|p{50mm}|}
        \hline
        \textbf{Threshold Factor} & \textbf{Unweighted Levenshtein} & \textbf{Weighted Levenshtein} \\
        \hline
        $5$ & ~~\llap{\textbullet}~~``This is an outlier'' \newline ~~\llap{\textbullet}~~``22nd of April 2004'' \newline ~~\llap{\textbullet}~~\textit{empty string} \newline ~~\llap{\textbullet}~~``30.09.2025'' & ~~\llap{\textbullet}~~``22nd of April 2004'' \\
        \hline
        $3$ & ~~\llap{\textbullet}~~``This is an outlier'' \newline ~~\llap{\textbullet}~~``22nd of April 2004'' \newline ~~\llap{\textbullet}~~\textit{empty string} \newline ~~\llap{\textbullet}~~``30.09.2025'' \newline ~~\llap{\textbullet}~~``1999-12-31'' \newline ~~\llap{\textbullet}~~``01042024'' \newline ~~\llap{\textbullet}~~``2099-99-99'' \newline ~~\llap{\textbullet}~~``0000-00-00'' \newline ~~\llap{\textbullet}~~``2012/01/01'' & ~~\llap{\textbullet}~~``22nd of April 2004'' \newline ~~\llap{\textbullet}~~\textit{empty string} \newline ~~\llap{\textbullet}~~``30.09.2025'' \newline ~~\llap{\textbullet}~~``2099-99-99'' \newline ~~\llap{\textbullet}~~``0000-00-00'' \newline ~~\llap{\textbullet}~~``1999-12-31'' \\
        \hline
        $2$ & ~~\llap{\textbullet}~~``This is an outlier'' \newline ~~\llap{\textbullet}~~``22nd of April 2004'' \newline ~~\llap{\textbullet}~~\textit{empty string} \newline ~~\llap{\textbullet}~~``30.09.2025'' \newline ~~\llap{\textbullet}~~``1999-12-31'' \newline ~~\llap{\textbullet}~~``01042024'' \newline ~~\llap{\textbullet}~~``2099-99-99'' \newline ~~\llap{\textbullet}~~``0000-00-00'' \newline ~~\llap{\textbullet}~~``2012/01/01'' & ~~\llap{\textbullet}~~``22nd of April 2004'' \newline ~~\llap{\textbullet}~~\textit{empty string} \newline ~~\llap{\textbullet}~~``30.09.2025'' \newline ~~\llap{\textbullet}~~``2099-99-99'' \newline ~~\llap{\textbullet}~~``0000-00-00'' \newline ~~\llap{\textbullet}~~``1999-12-31'' \newline ~~\llap{\textbullet}~~``01042024'' \newline ~~\llap{\textbullet}~~``2012/01/01'' \newline ~~\llap{\textbullet}~~``This is an outlier'' \\
        \hline
        $1.5$ & ~~\llap{\textbullet}~~``This is an outlier'' \newline ~~\llap{\textbullet}~~``22nd of April 2004'' \newline ~~\llap{\textbullet}~~\textit{empty string} \newline ~~\llap{\textbullet}~~``30.09.2025'' \newline ~~\llap{\textbullet}~~``1999-12-31'' \newline ~~\llap{\textbullet}~~``01042024'' \newline ~~\llap{\textbullet}~~``2099-99-99'' \newline ~~\llap{\textbullet}~~``0000-00-00'' \newline ~~\llap{\textbullet}~~``2012/01/01'' \newline ~~\llap{\textbullet}~~``2000-01-01'' & ~~\llap{\textbullet}~~``22nd of April 2004'' \newline ~~\llap{\textbullet}~~\textit{empty string} \newline ~~\llap{\textbullet}~~``30.09.2025'' \newline ~~\llap{\textbullet}~~``2099-99-99'' \newline ~~\llap{\textbullet}~~``0000-00-00'' \newline ~~\llap{\textbullet}~~``1999-12-31'' \newline ~~\llap{\textbullet}~~``01042024'' \newline ~~\llap{\textbullet}~~``2012/01/01'' \newline ~~\llap{\textbullet}~~``This is an outlier'' \newline ~~\llap{\textbullet}~~``2000-01-01''
        \\
        \hline
    \end{tabular}

    \caption{Detected outliers in the modified dataset for different thresholds}
    \label{tab:03_results_outlier_detection_modified}
\end{table}

    \chapter{Regular Expression-Based Approach} \label{chapter:04_regex_based_approach}
    Our second algorithm builds upon inferring regular expressions from a set of strings. We base the algorithm strategy on the assumption that there is a regular expression that describes the language of the expected data in the dataset. We can classify any string as outlier or normal data with the regular expression, depending on whether the regular expression matches it. We expand on the iterative Hierarchical Left Regular Expression (HiLRE) learning algorithm presented in \cite{doskoc_efficient_2016}, which we explain in the following section.

\section{Incremental Learning of Hierarchical Left Regular Expressions}
Researchers in \cite{doskoc_efficient_2016} introduce two essential concepts to infer a regular expression from a set of strings.
\begin{enumerate}[(1)]
    \item The concept of Hierarchical Left Regular Expressions (HiLRE), a restricted version of regular expressions, and 
    \item a learning algorithm, which infers an HILRE from the given dataset based on a hierarchical partition like the one we have proposed in \Cref{fig:03_proposed_hierarchy}.
\end{enumerate}

Hierarchical Left Regular Expressions are a restricted version of regular expressions, which results in a couple of crucial properties.
\begin{itemize}
    \item They are unambiguously parsable,
    \item regular expressions that are HiLREs do not have two following star-annotated elements, and
    \item an element of a HiLRE is never higher in the hierarchy than its following element.
\end{itemize}

Hierarchical Left Regular Expressions are defined as follows.

\begin{definition}[Hierarchical Left Regular Expressions (HiLRE)]
    Let $\Sigma$ be an alphabet and $P$ be a hierarchical partitioning thereof (see \Cref{def:03_hierarchical_partitioning}). Let $p_1,\dots,p_n\in P$. Then $r=r_1r_2\dots r_n$ is a \emph{HiLRE} if $r_1=p_1,r_n=p_n$ and for all $i$ with $1<i<n$ either
    \begin{enumerate}
        \item $r_i=p_i$, such that $p_i\not\supset p_{i+1}$, or
        \item $r_i=p_i^*$ such that $p_{i-1}\subseteq p_i$ and $p_i\bot p_{i+1}$
    \end{enumerate}
    The operator $\bot$ in this context means that $p_i$ is not a subset of $p_{i+1}$ and vice-versa, so $p_i\not\subset p_{i+1}$ and $p_{i+1}\not\subset p_i$.
\end{definition}

The HiLRE learning algorithm, as shown in the paper, works incrementally by keeping track of so-called Learnings, from which it can infer the resulting regular expression. The algorithm uses two sub-algorithms: The \texttt{initial\_string} algorithm, which provides an initial learning set for exactly one string, and the \texttt{add\_string} algorithm, which takes an existing list of learnings and modifies them, such that their resulting regular expression also matches the newly added string with a minimal HiLRE \cite{doskoc_efficient_2016}. That together means that for every step when inferring a HiLRE from the dataset, we can generate a regular expression that matches all strings considered up to that point.

A singular ``Learning'' thereby has four attributes to track which HiLRE element it currently represents, and which one it might represent later on when adding a new string using \texttt{add\_string}.
\begin{itemize}
    \item \texttt{element}, an element of the hierarchy that this Learning represents, for example ``$2$'', ``${0\text -9}$'' or ``$\Sigma$'' for the hierarchy $H$ given in \Cref{fig:03_proposed_hierarchy}.
    \item \texttt{max}, an intermediate calculation attribute used by the \texttt{add\_string} algorithm.
    \item \texttt{count}, for every ancestor to the element: how many subsequent characters does the ancestor match for the currently considered strings. The count monotonically increases for every higher component of the hierarchy, as a higher element matches the same number of characters or more.
    \item \texttt{has\_multiple}, for every ancestor to the element, as additional information to \texttt{count}, whether this element may match arbitrarily more characters than the count value. If \texttt{has\_multiple} is true, the Learning for this ancestor is equivalent to the star class in regular expression, provided that the \texttt{count} value for that ancestor is zero. Alternatively, for \texttt{count} values above zero, it is equivalent to the plus class.
\end{itemize}

Given a linked list of Learnings, the algorithm \texttt{generate\_reg\_ex} as proposed in \cite{doskoc_efficient_2016} infers a regular expression, a minimal HiLRE, which matches all learned strings.

To learn a HiLRE on a complete dataset without any outliers, one would run the \texttt{hilre\_generalize} algorithm. The algorithm first runs \texttt{initial\_string} on the first string to create an initial linked list of Learnings. Then it uses \texttt{add\_string} to add all the other strings from the dataset to the linked list of Learnings. Lastly, it returns a complete HiLRE for the dataset using \texttt{generate\_reg\_ex}.

\section{Finding Outliers Using the HiLRE Learning Algorithms}

The algorithm \texttt{hilre\_generalize} assumes that the dataset given does not contain any outliers. If it includes an outlier, the resulting HiLRE would also match it, providing a sub-par regular expression for the dataset. To accommodate outliers in the string dataset and to detect them successfully, we need a different approach based on the following assumption about the dataset:

\begin{proposition}
    \label{pro:4_2}
    There exists a HiLRE $H^*$ for the dataset $D$, such that every datum $d\in D$ with $d\in H^*$ is not an outlier, and every $d\in D$ with $d\notin H^*$ is an outlier. In short, $H^*$ is the language of all non-outliers, and $\overline {H^*}$ is the language of all outliers, respectively.
\end{proposition}

If the anomalous values get detected by $H^*$ too, the algorithm cannot find them, per \Cref{pro:4_2}. Examples of outliers, which the algorithm would be unable to detect, are ``$2012\text - 12 \text - 32$'', ``$2012\text - 13 \text - 99$'', or ``$2012\text - 02\text - 30$''. They are all invalid dates, and thus outliers on an ISO 8601 string set, but a regular expression with the given hierarchy can not distinguish them from valid dates. In the case of the first two examples, one could change the hierarchy to aid the algorithm in creating a better $H^*$ regular expression. One could think about adding $[0\text -1]$, $[0\text - 2]$, and $[0\text -3]$ hierarchical classes, so the HiLRE can be more restrictive in representing a valid date. However, for the last example, the algorithm needs the information that February has at most 29 days, and such a date string is therefore invalid. As discussed in \Cref{chapter:02_string_data}, using additional context is semantical outlier detection and thus out of scope for this thesis.

Our goal with the outlier detection algorithm is to find a HiLRE $H^*$ to categorize every string in the dataset as an outlier or an expected value. Before we introduce the algorithm to find $H^*$, we need to be able to detect whether one HiLRE is a subset of or equal to another HiLRE. 

Whether one regular expression includes another is computable for various forms of regular expression, including ones that are unambiguously parsable, as described in \cite{hovland_inclusion_2010}. As determined by \cite{doskoc_efficient_2016}, HiLREs fall under that category, meaning we can compute whether one HiLRE includes another HiLRE. We propose \Cref{alg:04_HiLRE_subseteq} to determine just that. For simplicity, we assume a HiLRE is a linked list containing HiLRE elements with the following properties.

\begin{enumerate}[(1)]
    \item \texttt{element}, the element of the hierarchy, this part of the regular expression represents,
    \item \texttt{count}, how many times \texttt{element} needs to match, and
    \item \texttt{has\_multiple}, whether more than \texttt{count} matches are allowed.
\end{enumerate}

\Cref{alg:04_HiLRE_subseteq} iterates element by element through the HiLRE lists, and checks whether the current hierarchical element of $h_1$ equals or is below the current hierarchical element of $h_2$. It advances to the next element of each HiLRE list when it reaches the required number of matches. To accommodate elements that allow for an arbitrary number of matches, the algorithm advances the element of the other HiLRE for every element in $h_1$ that gets matched by $h_2$. At the end, $h_1$ is a subset of or equal to $h_2$, if the algorithm wholly iterated through both HiLREs and found no invalid element. Notation-wise, $i_1$ and $i_2$ represent the index of the current element in $h_1$ or $h_2$, while $c_1$ and $c_2$ count the matched elements for $h_1$ and $h_2$.

Using the $\subseteq$ operator for two HiLREs as defined in \Cref{alg:04_HiLRE_subseteq}, we can now find a HiLRE $H^*$, which matches all non-outlier strings. We employ the following strategy. Out of all possible HiLREs for subsets of the dataset, we choose the one HiLRE that matches the most data points compared to the possible HiLREs, which are a subset of it.

\begin{figure}[hb]
    \centering
    \scalebox{0.84}{
        \begin{tikzpicture}[
            > = stealth,
            HiLRE/.append style = {
                draw = black,
                shape = rectangle,
                inner sep = 3pt
            },
            every path/.append style = {
                arrows = ->
            }]
            
            \node[HiLRE] (all) at (6, 7) {$[a\text -zA\text -Z0\text -9], 4$};
            \node[HiLRE] (alpha) at (3, 4) [fill=green] {$[a\text -z], 3$};
            \node[HiLRE] (numeric) at (12, 4) {$[0\text -9], 1$};
            \node[HiLRE] (a) at (-1, 2) {$a, 1$};
            \node[HiLRE] (b) at (3, 2) {$b, 1$};
            \node[HiLRE] (c) at (8, 2){$c, 1$};
            \node[HiLRE] (0) at (12, 2) {$0, 1$};
            \node[HiLRE] (none) at (6, -1) {$\emptyset, 0$};
            \path (all) edge node[] {$1$} (alpha);
            \path (all) edge node[above] {$3$} (numeric);
            \path (all) edge node[above] {$3$} (a);
            \path (all) edge node[right] {$3$} (b);
            \path (all) edge node[right] {$3$} (c);
            \path (all) edge node[below] {$3$} (0);
            \path (all) edge node[above left] {$4$} (none);
            \path (alpha) edge node[below] {$2$} (a);
            \path (alpha) edge node[left] {$2$} (b);
            \path (alpha) edge node[left] {$2$} (c);
            \path (alpha) edge node[left] {$3$} (none);
            \path (numeric) edge node[right] {$0$} (0);
            \path (numeric) edge node[right] {$1$} (none);
            \path (a) edge node[below] {$1$} (none);
            \path (b) edge node[left] {$1$} (none);
            \path (c) edge node[above] {$1$} (none);
            \path (0) edge node[below] {$1$} (none);
        \end{tikzpicture}
    }
    \caption{Example directed acyclic graph for all possible HiLREs on the dataset $D_4$}
    \label{fig:04-example-hilre-outlier}
\end{figure}
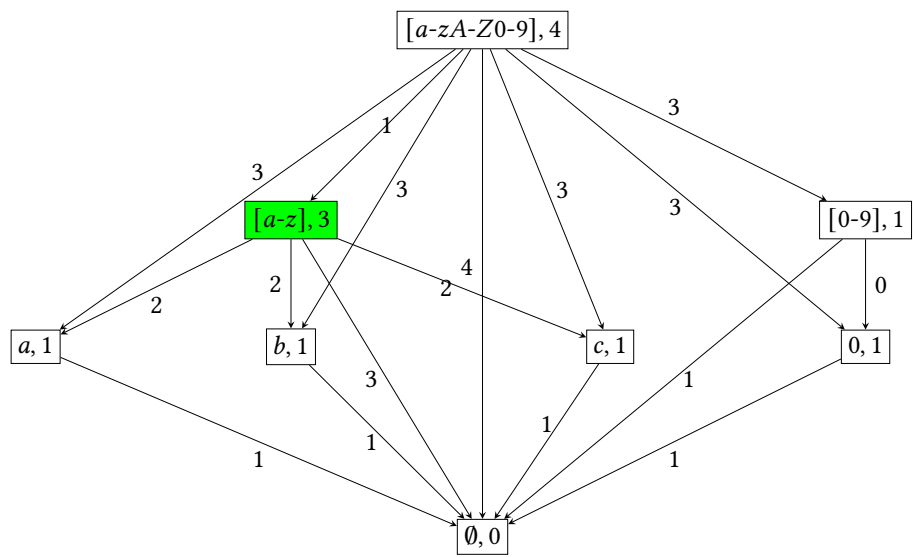

\begin{example}[Strategy to Find $H^*$]
    As an example for this strategy, we are looking at the dataset $D_4=\{a,b,c,0\}$. For every subset of the dataset, including the empty set, we can create the following HiLREs based on the hierarchy $H$ in \Cref{fig:03_proposed_hierarchy}, where every HiLRE matches every element of its corresponding subset. \[H_4=\{\emptyset, a, b, c, 0, [a\text -z], [0\text -9], [a\text -zA\text -Z0\text -9]\}\] From now on, $\emptyset$ represents the regular expression which matches nothing. See \Cref{fig:04-example-hilre-outlier} for a visualization of our strategy. It contains a directed acyclic graph, where every node represents a HiLRE and every edge represents whether the incoming HiLRE node is a subset of the outgoing HiLRE node. For explanatory purposes, every node contains the number of strings from the dataset that the HiLRE matches, and every path shows the difference in the number of matches between the two nodes it connects.

    We now choose the non-empty HiLRE with the largest minimal difference to all subset HiLREs as our $H^*$. In our example \Cref{fig:04-example-hilre-outlier}, we select the HiLRE node with the largest minimal outgoing edge weight. The minimal edge weights for this example are as follows.
    \begin{itemize}
        \item $[a\text -zA\text -Z0\text -9]$: $\min(1,3,3,3,3,3,4) = 1$
        \item $[a\text -z]$: $\min(2,2,2,3)=2$
        \item $[0\text -9]$: $\min(0,1)=0$
        \item $0, a, b, c$: $\min(1)=1$
    \end{itemize}
    Our chosen HiLRE $H^*$ to match non-outlier strings is $[a\text -z]$, which matches $a,b,c$ but not $0$. Therefore, for this dataset $D_4$, the algorithm would label $0$ as the only outlier.
\end{example}

\begin{algorithm}
    \caption{Is a HILRE $h_1$ a subset of or equal to another HiLRE $h_2$ ($h_1\subseteq h_2$)?}
    \label{alg:04_HiLRE_subseteq}
    \KwData{HiLRE $h_1$ and $h_2$, Hierarchy $H$}
    \KwResult{Boolean $b$}
    \If{$|h_1| = 0$}{\Return True\;}
    $i_1 \gets 0$\;
    $i_2 \gets 0$\;
    $c_1 \gets 0$\;
    $c_2 \gets 0$\;
    \While{$i_1 < |h_1| \land i_2 < |h_2|$}{
        \While{$c_1 < h_1[i_1].\operatorname{count} \land c_2 < h_2[i_2].\operatorname{count}$}{
            \If{$h_1[i_1].\operatorname{element} \not\subseteq_H h_2[i_2].\operatorname{element}$}{
                \Return False\;
            }
            $c_1 \gets c_1 + 1$\;
            $c_2 \gets c_2 + 1$\;
        }
        
        \If{$c_2 = h_2[i_2].\operatorname{count}$}{
            \If{$h_2[i_2].\operatorname{has\_multiple}$}{
                \If{$c_1 = h_1[i_1].\operatorname{count}$}{
                    $i_1 \gets i_1 + 1$\;
                    $c_1 \gets 0$\;
                }
                \While{$i_1 < |h_1| \land h_1[i_1].\operatorname{element} \subseteq_H h_2[i_2].\operatorname{element}$}{
                    $i_1 \gets i_1 + 1$\;
                    $c_1 \gets 0$\;
                }
                $i_2 \gets i_2 + 1$\;
                $c_2 \gets 0$\;
                \textbf{continue}\;
            }
            $i_2 \gets i_2 + 1$\;
            $c_2 \gets 0$\;
        }
        
        \If{$c_1 = h_1[i_1].\operatorname{count}$}{
            \If{$h_1[i_1].\operatorname{has\_multiple}$}{
                \While{$i_2 < |h_2| \land h_1[i_1].\operatorname{element} \subseteq_H h_2[i_2].\operatorname{element}$}{
                    $i_2 \gets i_2 + 1$\;
                    $c_2 \gets 0$\;
                }
            }
            $i_1 \gets i_1 + 1$\;
            $c_1 \gets 0$\;
        }
    }
    \Return $i_1 = |h_1| \land i_2 = |h_2|$\;
\end{algorithm}

Overall, our HiLRE outlier detection algorithm consists of the following three parts.
\begin{enumerate}[(1)]
    \item Retrieve the HiLREs for all possible subsets of the dataset,
    \item for every HiLRE, calculate the minimal difference to all subset HiLREs, and then choose the HiLRE with the maximum minimal difference as $H^*$, and
    \item match all strings from the dataset against $H^*$, labeling non-matching strings as outliers.
\end{enumerate}

To retrieve all possible HiLREs, we propose \Cref{alg:04_all_HiLREs} \texttt{get\_all\_hilres}. It utilizes a set of candidate HiLREs $C$, which themselves are in the form of linked lists of Learnings. The algorithm initializes this set $C$ with the initial Learnings of all strings in the dataset, calculated by \texttt{initial\_string}. Then, we iterate through every candidate HiLRE in our set, and iteratively add every possible non-matching string from the dataset. We add the resulting Learnings to our candidate set $C$ if it does not include them yet. In the end, $C$ contains every HiLRE of all possible subsets of $D$.

\begin{algorithm}
    \caption{Retrieve HiLREs for all possible subsets of the dataset (\texttt{get\_all\_hilres}).}
    \label{alg:04_all_HiLREs}
    \KwData{Set of strings $D$, hierarchy $H$}
    \KwResult{Set of candidate HiLREs in the form of linked lists of Learnings $C$}
    $C \gets \{\}$\;
    \For{$s\in D$}{
        $C \gets C\cup\{\text{\texttt{initial\_string}}(s)\}$\;
    }
    \For{$c\in C$}{
        \For{$s\in D$}{
            \If{$s\notin c$}{
                $c_{\text{new}}\gets \texttt{add\_string}(c,s)$\;
                \If{$c_{\text{new}}\notin C$}{
                    $C\gets C\cup\{c_{\text{new}}\}$\;
                }
            }
        }
    }
    \Return $C$\;
\end{algorithm}

To select the HiLRE $H^*$ from all possible HiLREs, we propose \Cref{alg:04_find_H_star} \texttt{find\_H\_star}. It iterates through the candidate HiLREs $C$ from $\texttt{get\_all\_hilres}$, ordered by the number of matches. In case of the explanatory acyclic directed graph in \Cref{fig:04-example-hilre-outlier}, we would iterate through the HiLRE nodes from bottom to top. For every candidate HiLRE $c\in C$, we retrieve the set $C_\subset$. It contains all candidate HiLREs, which are a subset of $c$. We then calculate the difference in the number of string matches for every subset candidate $c'\in C_\subset$ and select the minimum as $d$. If $d$ is larger than the previously saved maximum $d_{\max}$, we update $H^*$ to the current candidate $c$. After we reviewed all candidates, $H^*$ contains the HiLRE to match the strings against.

\begin{algorithm}
    \caption{Select HiLRE $H^*$ from all possible HiLREs (\texttt{find\_H\_star}).}
    \label{alg:04_find_H_star}
    \KwData{Dataset of strings $D$, hierarchy $H$, set of HiLREs in the form of linked lists of Learnings $C$ ordered by number of matches}
    \KwResult{HiLRE $H^*$ in the form of a linked list of Learnings}
    $H^*\gets \emptyset$\;
    $d_{\max}\gets 0$\;
    \For{$c\in C$}{
        $C_\subset\gets \{c'~|~c'\in C\land c'\subset c\}$\;
        $n\gets\#\{s~|~s\in D\land s\in c\}$\;
        $d\gets \min\left\{n-\#\{s~|~s\in D\land s\in c'\}~|~c'\in C_\subset\right\}$\;
        \If{$d>d_{\max}$}{
            $H^*\gets c$\;
            $d_{\max}\gets d$\;
        }
    }
    \Return $H^*$\;
\end{algorithm}

The complete outlier detection algorithm described in \Cref{alg:04_detect_HiLRE_outlier} combines all these algorithms. It first retrieves all candidate HiLREs using \texttt{get\_all\_hilres}, then finds $H^*$ using \texttt{find\_H\_star}. Lastly, it creates a set of outlier strings with all strings that are not matched by $H^*$.

\begin{algorithm}
    \caption{Outlier detection using HiLREs.}
    \label{alg:04_detect_HiLRE_outlier}
    \KwData{Dataset of string $D$, hierarchy $H$}
    \KwResult{Set of outlier strings $O\subset D$}
    $C\gets \texttt{get\_all\_hilres}(D,H)$\;
    $\texttt{sort}(C)$\;
    $H^*\gets \texttt{find\_H\_star}(D,H,C)$\;
    $O\gets\{s~|~s\in D\land s\notin H^*\}$\;
    \Return $O$\;
\end{algorithm}

\subsection{Variant With a Parameter for the Minimum Number of Matches}
The default variant of the algorithm presented above always selects the HiLRE with the most new matches compared to its subset HiLREs. However, this is not always the ideal HiLRE, as the following example shows.

\begin{example}[Non-ideal HiLRE Selection]
    Take a dataset with ten ``2020-01-01'' strings, nine strings with the dates following on ``2020-01-01'', and a string ``This is an outlier''. Ideally, the algorithm should detect the string ``This is an outlier'' as the outlier. However, as the HiLRE for ``2020-01-01'' has the most change to its only subset HiLRE $\emptyset$, the algorithm selects it as the HiLRE for $H^*$. Therefore, all strings different from ``2020-01-01'' get classified as outliers.
\end{example}

To circumvent this behavior and enhance our ability to adjust the algorithm to the dataset, we propose a parameter $p_{\min}\in\mathbb R$ with $0\le p_{\min}\le 1$. It introduces another condition to select the HiLRE $H^*$, as it must match at least $p_{\min}\cdot |D|$ strings from the dataset $D$. With $p_{\min}=0$, this variant yields the same results as the default variant outlined above. See \Cref{alg:04_find_H_star_variant} for a variant of the \texttt{find\_H\_star} algorithm, which implements this parameter. Changes to the algorithm are in bold text.

\begin{algorithm}
    \caption{Variant to select HiLRE $H^*$ from all possible HiLREs (\texttt{find\_H\_star\_variant}).}
    \label{alg:04_find_H_star_variant}
    \KwData{Dataset of strings $D$, hierarchy $H$, set of HiLREs in the form of linked lists of Learnings $C$ ordered by number of matches, \textbf{minimum ratio of matches $p_{\min}\in\mathbb R,0\le p_{\min}\le 1$}}
    \KwResult{HiLRE $H^*$ in the form of a linked list of Learnings}
    $H^*\gets \emptyset$\;
    $d_{\max}\gets 0$\;
    \For{$c\in C$}{
        $C_\subset\gets \{c'~|~c'\in C\land c'\subset c\}$\;
        $n\gets\#\{s~|~s\in D\land s\in c\}$\;
        $d\gets \min\left\{n-\#\{s~|~s\in D\land s\in c'\}~|~c'\in C_\subset\right\}$\;
        \If{$d>d_{\max}\mathbf{\land n\ge p_{\min}\cdot |D|}$}{
            $H^*\gets c$\;
            $d_{\max}\gets d$\;
        }
    }
    \Return $H^*$\;
\end{algorithm}

\begin{example}[Non-ideal HiLRE Selection Example With the New Parameter]
    With this algorithm variant, we can specify a value $p_{\min}$ for the minimum number of strings the HiLRE $H^*$ must match. If we set $p_{\min}=0.75$, the algorithm selects ``2020-01-0[0-9]'' as the HiLRE for the sample dataset $D_{e}$ mentioned in the above example. The algorithm then labels the strings ``2020-01-10'' and ``This is an outlier'' as outliers. If we want to narrow the outlier set even more, we can set $p_{\min}$ to $0.95$. This way, the algorithm selects ``2020-01-[0-9]\{2\}'' as the HiLRE for $H^*$, and only labels ``This is an outlier'' as an outlier.
\end{example}

\section{Tests on Synthetic Data}

To test our HiLRE-based outlier detection algorithm, we use the same two datasets as in \Cref{chapter:03_k_nn_levenshtein}. The first is a set of a thousand consecutive ISO 8601 strings, starting from ``2020-01-01''. The second is a changed version of the first set, where we have replaced ten random values with outliers of varying degrees as listed in \Cref{tab:03_new_values_in_the_data_set}.

We run two variants on those two datasets. First, the default version of the algorithm, and second, the modified version with $p_{\min}=0.85$. As results for the algorithm runs, we provide the selected HiLRE $H^*$ and the corresponding set of outliers.

Both algorithms select the same HiLRE as $H^*$ for the modified and unmodified datasets. Which HiLRE they choose for both datasets is different between the two algorithms, though. The default variant selects ``202[0-9]-0[0-9]-[0-9]\{2\}'' as $H^*$ for both datasets, so it classifies all dates in October, November, and December as outliers. This result is to be expected, as there are far more dates with a zero instead of a one as the first digit for the month. The algorithm variant with $p_{\min}=0.85$ selects a different $H^*$ for both dataset, namely ``202[0-9]-[0-9]\{2\}-[0-9]\{2\}''. This HiLRE does not yield any false positives and successfully detects all ten outliers on the modified dataset.

In the following \Cref{chapter:05_using_real_world_data}, we investigate this algorithm deeper and compare it to the $k$-nearest neighbor algorithm given in \Cref{chapter:03_k_nn_levenshtein}.

    \chapter{Comparison and Experiments} \label{chapter:05_using_real_world_data}
    In this chapter, we compare the local outlier factor (LOF) outlier detection algorithm presented in \Cref{chapter:03_k_nn_levenshtein} to the HiLRE-based outlier detection algorithm presented in \Cref{chapter:04_regex_based_approach}. In the first section, we explain the methodology of the experiments and introduce the datasets we use. In the section after that, we run experiments on the clean datasets to get a baseline result and evaluate their false positive rate on datasets with supposedly no outliers. Lastly, we compile multiple different datasets to assess the algorithm's ability to detect one or multiple classes of outliers.

\section{Methodology}
A single experiment uses one algorithm and one dataset specific to it. The dataset consists of a base dataset $D$ with expected values and several outlier datasets $O_i$, from which we each take $k_i\in\mathbb N$ with $1\le k_i\le |O_i|$ values. For an experiment with $n\in \mathbb N$ values, we thus have $o=\sum_{i=1}^{|O|} k_i$ total outliers, and it must hold that $o\le n$. We randomly choose $n$ normal values from the base dataset $D$ to create the simulation-specific dataset. Then, we choose $k_i$ random values for every selected outlier dataset and replace $o$ normal values in the dataset with them. This way, we get a dataset with $n$ total values, from which $\frac{o}{n}$ percent are outliers. As we randomly sample data from our datasets, it could happen that one sampled dataset randomly benefits one of the two algorithms. Therefore, we run every simulation configuration $100$ times, each with different randomly sampled values from the underlying datasets. As the simulation result, we then take the average detected outlier rate and false positive rate, and compare them to the other simulations.

We run every simulation with a set of different parameter values. With the LOF algorithm, we use \[P_{\operatorname{LOF}}=\left\{\frac i{20}\mid i\in\mathbb N\land 20\le n\le 100\right\}=\{1, 1.05, 1.1, \dots 5\}\] as the values for the threshold factor, and for the HiLRE algorithm we use \[P_{\operatorname{HiLRE}}=\left\{\frac{i}{20}~\mid i\in\mathbb N\land 0\le i\le 20\right\}=\{0,0.05,0.1,\dots 1\}\] as percentage values for the minimum number of strings $p_{\min}$ the $H^*$ HiLRE needs to match.

The real-world datasets we use stem from the quality reports of all German hospitals, published by the Federal Joint Committee \cite{gemeinsamer_bundesausschuss_qualitatsberichte_2025}. In particular, the data for the experiments is from the year 2023, and we only use the address strings (zip-codes, county names, street names, and house numbers), date, and time strings from the hospitals' contact fields. We used \cite{radiker_xmloot_2025} to extract the data into a suitable format. One can access the code with which we run the experiments and achieve the presented results online, see \cite{maus_code_2025}. It also includes the code to reproduce the results for \Cref{chapter:03_k_nn_levenshtein} and \Cref{chapter:04_regex_based_approach}, as well as further experiments and their results.

\section{Datasets Without Outliers}

In this section, we evaluate the algorithm on the unmodified datasets to compare their number of false positives over different parameter settings. We run the algorithm on all of our datasets, each with $n=1000$ total elements. In this case, the number of false positives amounts to the number of detected outliers, as the datasets do not contain any outliers. Therefore, we ideally expect zero detected outliers and thus zero false positives. To visualize our result, we create a plot for each algorithm, where every line represents the ratio of false positives to the total number of strings for one dataset over the parameter values of the algorithm. 

\begin{figure}
    \centering
    \includegraphics[width=0.9\linewidth]{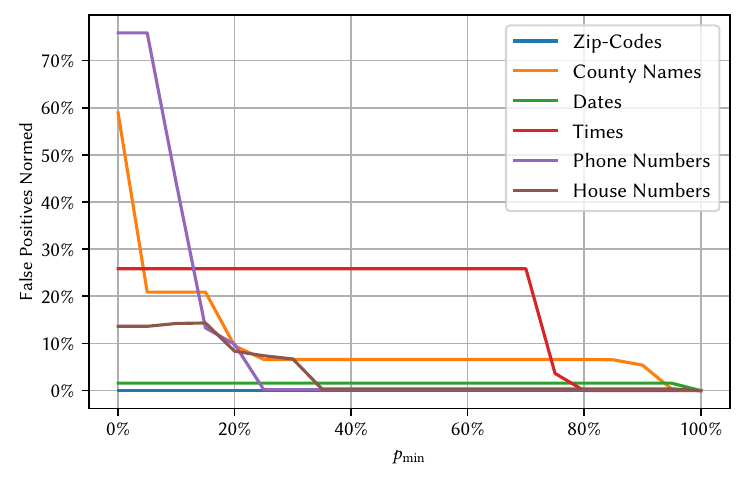}
    \caption{False positives for the HiLRE-based algorithm on a selection of datasets}
    \label{fig:05_clean_hilre_normed}
\end{figure}

\Cref{fig:05_clean_hilre_normed} shows the results for the HiLRE-based algorithm. At a parameter value over $p_{\min}=35\%$, the algorithm reaches zero false positives for most datasets, except the date, time, and county names dataset. For the zip-code dataset, it immediately selects the correct regular expression, namely exactly five digits.
The result for the date set stems from a bias in the date dataset. As most hospitals submit their quality report towards the end of the year, the HiLRE-based algorithm selects a regular expression, where the first digit of the month must be a $1$. Only for a $p_{\min}$ over $95\%$, the algorithm manages to select a more general date regular expression and thus detect no strings as outliers. The time dataset has a similar problem. As most quality reports get submitted during standard work times, the selected regular expression demands that the hour also start with a one. There, a higher $p_{\min}$ also leads to a more general time regular expression and thus no false positives anymore. The county names are very diverse, so it is hard to find a closely fitting regular expression to describe them. For example, it contains short strings consisting only of letters such as ``Bonn'', but also longer strings with different characters such as ``Frankfurt (Oder)''. With a higher $p_{\min}$, it finds a regular expression that encompasses all county names. However, as the regular expression only describes the data quite loosely, it may also classify possible outliers as expected data, which we see in a later section.

\begin{figure}
    \centering
    \includegraphics[width=0.9\linewidth]{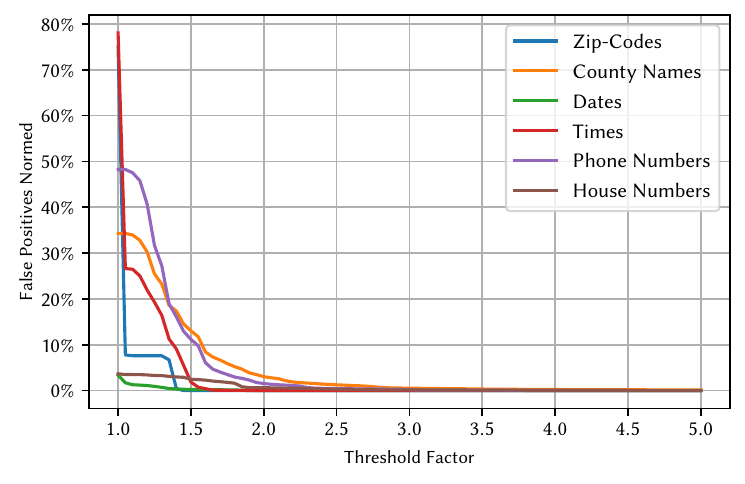}
    \caption{False positives for the LOF algorithm with the default Levenshtein metric on a selection of datasets}
    \label{fig:05_clean_lof_unweighted_normed}
\end{figure}

\begin{figure}
    \centering
    \includegraphics[width=0.9\linewidth]{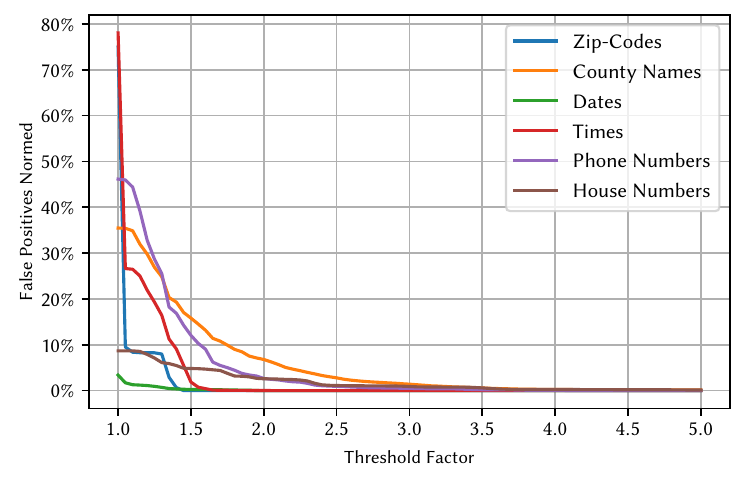}
    \caption{False positives for the LOF algorithm with the hierarchically weighted Levenshtein metric on a selection of datasets}
    \label{fig:05_clean_lof_weighted_normed}
\end{figure}

\Cref{fig:05_clean_lof_unweighted_normed} and \Cref{fig:05_clean_lof_weighted_normed} show the false positives for the two LOF algorithms. Both rates show a high false positive score at the lowest threshold, which rapidly decreases with a threshold value up to $1.5$. After that, the shrinkage slows down until the false positives eventually reach zero. The phone number and the county names datasets reach zero with a higher threshold value, as they contain some strings whose lengths deviate largely from their average length. Therefore, their distance to the other data is larger, and we require a larger threshold to exclude them from the outlier list. 

The LOF algorithm using the hierarchical Levenshtein measure proposed in \Cref{sec:03_distance_measures_for_strings}  shows a similar result to the LOF algorithm that uses the default Levenshtein metric, because we are working with clean datasets. As the structure of the strings within one dataset is similar, especially with the zip-codes, the time, and the date strings, the hierarchical weights do not come into effect. We only see a difference with datasets, where some strings have a different structure than others. Take the house number dataset, for example. There are some house numbers, which contain letters, such as ``Gebäude 3''. This string is sufficiently different from most other strings in this dataset, which consist only of numbers. Therefore, we need a higher threshold value to reach zero false positives for this dataset. 

In conclusion, one can find a parameter high enough for all algorithms where the number of false positives on a clean dataset reaches zero. The number of false positives for smaller thresholds heavily depends on the dataset and algorithm. The LOF algorithm with the unweighted distance measure reaches zero false positives with a lower threshold. In contrast, the LOF algorithm with the hierarchically weighted distance measure allows for a finer distinction between the data, as strings consisting of characters from different character classes are further apart. The number of false positives of the HiLRE algorithm heavily depends on how much structure is in the dataset and how distributed the different characters from a possible character class are in the strings. A higher value for $p_{\min}$ eventually leads to zero false positives, at least with $p_{\min}=1$. However, these settings might lead to a smaller true positive rate when using a dataset with outliers, which we examine in the following section.

\section{Datasets With Outliers}

This section looks at datasets with a certain percentage of outliers. For that, we select $n=1000$ data points from a dataset representing the normal data, and replace a percentage of these data with data from one or more other datasets that the algorithms should detect as outliers. We explain which concrete datasets we use later when we explain the concrete experiment. To evaluate the results, we make use of receiver operating characteristic (ROC) plots as described in \cite{fawcett_introduction_2006}. Our ROC plots are scatter plots, which contain one point for every simulation result. The x-axis shows the false positive rate, which is the percentage of false positives in the detected outlier set. At the same time, the y-axis displays the true positive rate, so the percentage of detected outliers. The diagonal of a ROC plot is equal to the performance of randomly selecting data points as outliers, so the further away from the diagonal towards the top left of the plot a result lies, the better it is. As we run the algorithms for a vast range of parameters, as explained at the beginning of the chapter, we select the parameter for each algorithm, which yields the best result towards the top left of the plot.

\subsection{Zip-Codes as Expected Data, County Names as Outliers}

The first dataset configuration we look at uses the zip-code dataset as the dataset for the normal data, while we use the county names dataset as the outlier dataset. We create different versions of the simulation dataset, where each dataset contains $1000$ total elements with $1,5,10,33$, or $50$ percent outliers from the county names set, and the rest being data from the zip-code dataset. 

\begin{figure}[hb]
    \centering
    \includegraphics[width=\lofwidth]{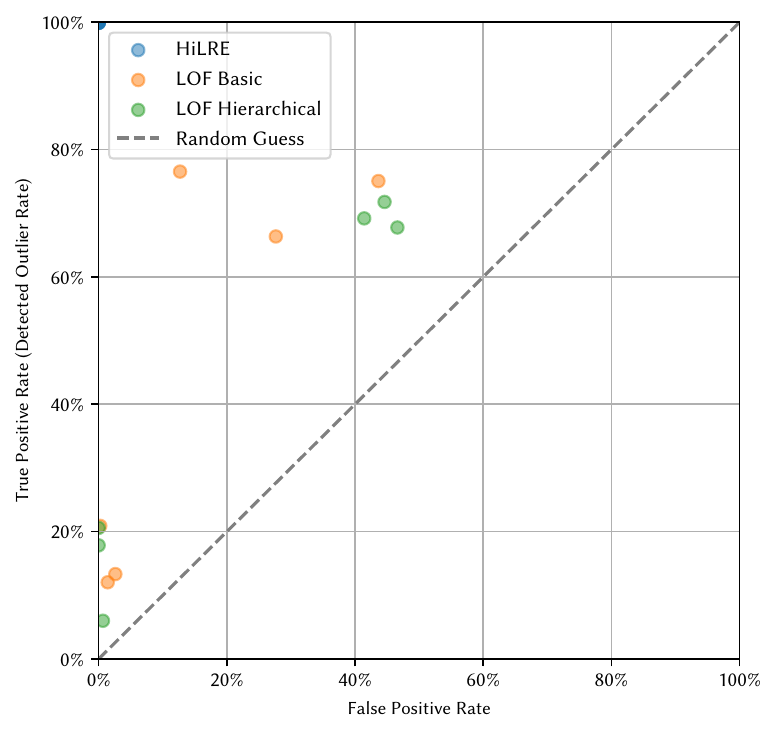}
    \caption{Percentage of detected outliers for the three algorithms using datasets based on the zip-code dataset with various percentages of county name outliers}
    \label{fig:05_plz_single_roc_chart}
\end{figure}

\Cref{fig:05_plz_single_roc_chart} shows the ROC plot for this dataset configuration. The HiLRE-based algorithm can correctly identify every outlier for every possible dataset configuration with zero false positives. The LOF algorithms are also able to detect a majority of the outliers, depending on the percentage of outliers in the dataset. Their results are better for datasets with fewer outliers, and worse for datasets with more outliers. Compared to the LOF algorithm with the standard Levenshtein distance metric, the hierarchically weighted variant yields more stable results at the cost of generally detecting fewer outliers.

The ROC curve shows that all algorithms are capable of detecting outliers in the dataset. The HiLRE-based algorithm performs exceptionally well because it successfully selects the regular expression that describes the five-digit zip-codes for every dataset configuration. The LOF algorithms sometimes group county names with the zip-codes if they have the same five-character length. That explains why the algorithm sometimes fails to detect these county names as outliers, or instead labels some zip-codes as outliers if every character in them is different from the majority of the other zip-codes, and the Levenshtein metric uses five replacement operations as the distance. The observation with the replacement operations also explains why the hierarchical variant is more stable, as it weights the replacement operations according to the given hierarchy and thus yields a higher distance between same-length county names and zip-codes than the default Levenshtein metric.

\subsection{County Names as Expected Data, Zip-Codes as Outliers}

For the next experiment, we swap the expected dataset and the outlier dataset from the previous experiment. This way, data from the county names dataset is the expected data, while the algorithms should label the data from the zip-codes dataset as outliers. This experiment contrasts with the previous experiment and shows the result if the dataset is the other way around.

\begin{figure}[ht]
    \centering
    \includegraphics[width=\lofwidth]{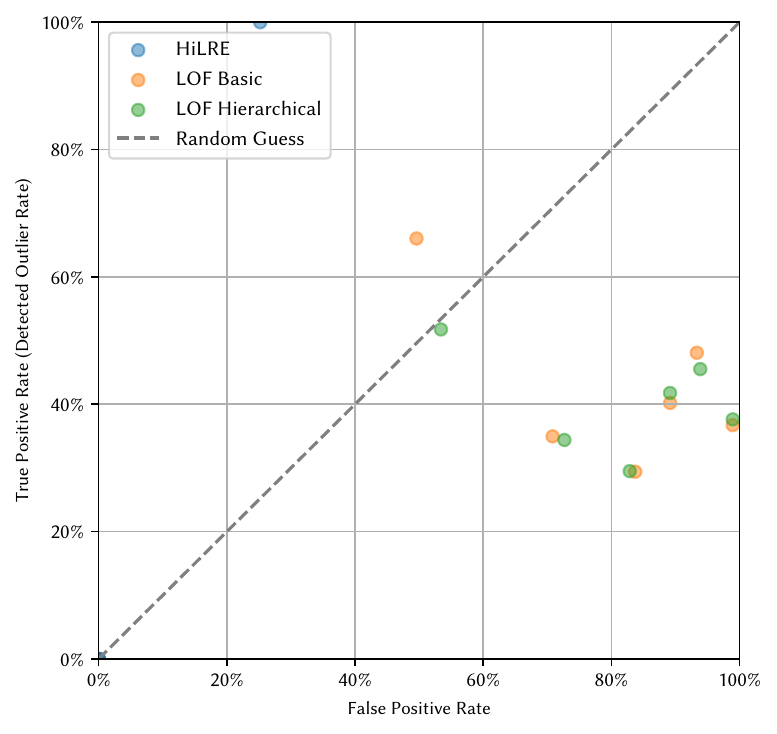}
    \caption{Percentage of detected outliers for the three algorithms using datasets based on the county names with various percentages of zip-code outliers}
    \label{fig:05_county_single_roc_chart}
\end{figure}

\Cref{fig:05_county_single_roc_chart} contains the ROC plot for this experimental setup. We can see that the HiLRE algorithm manages to detect some outliers on the dataset with only one percent of outliers, but fails to label any datum as an outlier for datasets containing more outliers. The LOF algorithms fail to reliably detect outliers on these datasets, as their best results are equal to or just slightly better than the random guesser, with the majority of their results containing a vast number of false positives with only a few true positives. 

The reason for this bad result is the nature of the dataset. The county names are pretty scattered. A regular expression can not describe them; they have vastly variable lengths and consist of different characters from various character classes in the hierarchy. Therefore, adding the zip-codes to the data introduces some noise into the dataset, which was also there before. The LOF algorithms can not distinguish between the noise that is already present in the dataset and the zip-code outliers we introduce. The HiLRE does manage to guess a rather good regular expression for the dataset with one percent outliers, but fails to do so for datasets with more outliers. Using a different selection mechanic for the $H^*$ regular expression could perhaps change this result.

\subsection{Zip-Codes as Expected Data, House and Phone Numbers as Outliers}

For the third experiment, we select data from the zip-code dataset as expected data, while the outliers stem from the house and phone numbers dataset. We use two dataset configurations. The first contains one percent of outliers of each outlier set, while the second one includes ten percent of outliers of each outlier dataset. These configurations allow us to evaluate the performance of the algorithm if the general structure of outliers and normal data is the same, but their length is different.

\begin{figure}[hb]
    \centering
    \includegraphics[width=\lofwidth]{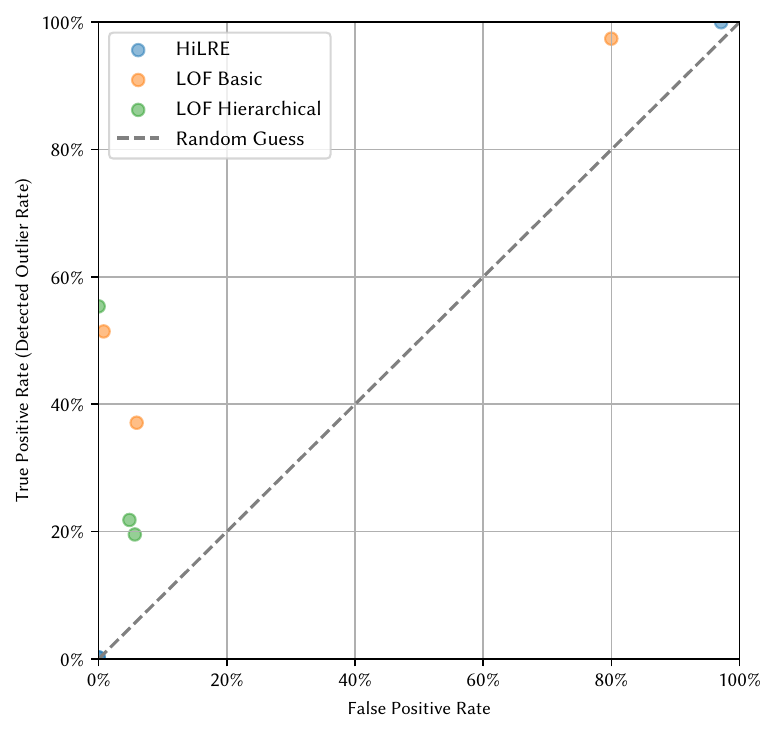}
    \caption{Percentage of detected outliers for the three algorithms using datasets based on the zip-code dataset with various percentages of house number and phone number outliers}
    \label{fig:05_numerical_tests_roc_chart}
\end{figure}

\Cref{fig:05_numerical_tests_roc_chart} shows the ROC plot for this experiment. The LOF algorithms do generally label the correct data as outliers, with a low false positive rate and a true positive rate of around fifty to sixty percent for the dataset with a total of twenty percent outliers and around twenty to forty percent for the datasets with fewer outliers. The exception for the LOF algorithms is the best result of the algorithm with the default Levenshtein metric for the dataset with a total of ten percent outliers. It has an exceptionally high false positive rate, while also detecting most outliers. The HiLRE-based algorithm struggles on this dataset. It detects almost all data as outliers for the dataset that has a total of ten percent outliers, while the best result for all the other datasets is the empty set.

The dataset mainly consists of digit strings with varying lengths. Therefore, the HiLRE-based algorithm struggles to find a correct HiLRE to identify the expected data, as it adds all the regular expressions to the pool of possible $H^*$ expressions that match a specific number \emph{or more} digits. Then, the algorithm either selects a regular expression that is too underfitting or too overfitting, which results in the observed results. On the other hand, the LOF algorithms perform well on these datasets, as the strings stemming from one dataset are closer to strings from the same dataset, as opposed to strings from other datasets.. As the zip-code cluster contains more data points and is denser than the others, the algorithm successfully detects many outliers. Some house numbers, which are longer and thus close to the zip-codes, do not get detected as outliers. The same holds for some shorter phone numbers, which explains the true positive rate not being as high as one would expect.

    \chapter{Conclusions and Outlook} \label{chapter:06_conclusions}
    In this thesis, we introduced two algorithms to detect syntactical outliers in single-word datasets and compared them. 

In \Cref{chapter:03_k_nn_levenshtein}, we introduced the first algorithm, the local outlier factor algorithm. We combined it with the Levenshtein distance measure to take string data instead of numerical data as input. We proposed an additional approach to weighting the Levenshtein measure, which utilizes a hierarchical partition of the incoming alphabet to represent potential structural elements of the strings. To enhance the algorithm, we introduced a different thresholding method, and to reduce the parameter inputs, we utilized the $KFCS$ guesser to choose a value for $k$. We experimentally showed that this algorithm can find density-based outliers on a synthetic string dataset.

In \Cref{chapter:04_regex_based_approach}, we introduced the second algorithm, built upon the hierarchical left regular expression learner. The learner used the same hierarchical partition of the incoming alphabet as the proposed weighted distance measure to create restricted regular expressions. Using the learner, the algorithm created hierarchical left regular expressions for all subsets of the data, and chose the one that detected the most values compared to all subset regular expressions. It then used this regular expression to classify all strings as outliers or non-outliers. To enhance the selection process, we introduced a variant of this algorithm that uses a parameter $p_{\min}$ to define the minimum percentage of data the selected regular expression needs to classify as non-outliers. We showed experimentally that this algorithm can find HiLRE-based outliers in the same synthetic dataset as with the first algorithm.

In \Cref{chapter:05_using_real_world_data}, we used datasets containing date strings, time stamps, and address information to compare the two algorithms. We compiled them together to create datasets that contain a fixed number of outliers of varying types. In the experiments on those datasets, we observed that the results of the outlier detection algorithms heavily depend on the dataset at hand. The HiLRE-based algorithm worked best if a hierarchical left regular expression could fit the expected data closely while not matching the outliers. This way, it could find many outliers with relatively few false positives and negatives. However, if no closely fitted hierarchical left regular expression could represent the expected data, the algorithm underperformed, sometimes detecting zero outliers. The local outlier factor algorithm did not create such extreme results, often finding many outliers at the cost of a higher false positive and false negative rates. It was best at detecting outliers if they consisted of the same general character set but varied in length compared to the expected data. The weighted distance provided a way to tailor the algorithm's result to the dataset, yielding sometimes better and sometimes worse results than the unweighted measure. 

\section{Future Work}
While we experimentally showed that the HiLRE algorithm can find outliers, we have not touched on theoretical analysis, such as time and space usage, or an analysis of the correctness of the algorithm. One could also consider using different criteria when selecting a HiLRE to classify the strings into outliers, as enhancing this measure could drastically improve the results of the HiLRE-based algorithm.

Generally, there is not much research concerning outlier detection on string data. With our experiments, we looked into detecting syntactical outliers on single-word strings. We did not touch on detecting outliers on strings that contain multiple words or require additional context to be classified as outliers. In our experiments, we sometimes discovered properties of the datasets with the outlier detection algorithms that we had not known about before. For example, we learned that most hospitals submit their quality reports towards the end of the year between $10$ and $20$ o'clock. Therefore, gaining insights into a dataset using outlier detection algorithms, especially the HiLRE-based outlier detection algorithm, could also be a topic of further investigation.

    \pagestyle{plain}

    \renewcommand*{\bibfont}{\small}
    \printbibheading
    \addcontentsline{toc}{chapter}{Bibliography}
    \printbibliography[heading = none]

@article{domingues_comparative_2018,
	title = {A comparative evaluation of outlier detection algorithms: Experiments and analyses},
	volume = {74},
	issn = {00313203},
	url = {https://linkinghub.elsevier.com/retrieve/pii/S0031320317303916},
	doi = {10.1016/j.patcog.2017.09.037},
	shorttitle = {A comparative evaluation of outlier detection algorithms},
	pages = {406--421},
	journaltitle = {Pattern Recognition},
	shortjournal = {Pattern Recognition},
	author = {Domingues, Rémi and Filippone, Maurizio and Michiardi, Pietro and Zouaoui, Jihane},
	urldate = {2025-08-30},
	date = {2018-02},
	langid = {english},
}

@incollection{hovland_inclusion_2010,
	location = {Berlin, Heidelberg},
	title = {The Inclusion Problem for Regular Expressions},
	isbn = {978-3-642-13088-5},
	url = {http://link.springer.com/10.1007/978-3-642-13089-2_26},
	pages = {309--320},
	booktitle = {Lecture Notes in Computer Science},
	publisher = {Springer Berlin Heidelberg},
	author = {Hovland, Dag},
	editor = {Dediu, Adrian-Horia and Fernau, Henning and Martín-Vide, Carlos},
	urldate = {2025-07-23},
	date = {2010},
	doi = {10.1007/978-3-642-13089-2_26},
	note = {{ISSN}: 0302-9743, 1611-3349},
}

@book{hawkins_identification_1980,
	location = {Dordrecht},
	title = {Identification of Outliers},
	rights = {http://www.springer.com/tdm},
	isbn = {978-94-015-3994-4},
	url = {http://link.springer.com/10.1007/978-94-015-3994-4},
	publisher = {Springer Netherlands},
	author = {Hawkins, D. M.},
	urldate = {2025-07-06},
	date = {1980},
	langid = {english},
	doi = {10.1007/978-94-015-3994-4},
}

@inproceedings{bakar_comparative_2006,
	location = {Bangkok, Thailand},
	title = {A Comparative Study for Outlier Detection Techniques in Data Mining},
	isbn = {978-1-4244-0022-5},
	url = {http://ieeexplore.ieee.org/document/4017846/},
	doi = {10.1109/ICCIS.2006.252287},
	eventtitle = {2006 {IEEE} Conference on Cybernetics and Intelligent Systems},
	pages = {1--6},
	booktitle = {2006 {IEEE} Conference on Cybernetics and Intelligent Systems},
	publisher = {{IEEE}},
	author = {Bakar, Zuriana and Mohemad, Rosmayati and Ahmad, Akbar and Deris, Mustafa},
	urldate = {2025-08-25},
	date = {2006-06},
}

@article{xu_comparison_2018,
	title = {A Comparison of Outlier Detection Techniques for High-Dimensional Data:},
	volume = {11},
	issn = {1875-6883},
	url = {https://link.springer.com/10.2991/ijcis.11.1.50},
	doi = {10.2991/ijcis.11.1.50},
	shorttitle = {A Comparison of Outlier Detection Techniques for High-Dimensional Data},
	pages = {652},
	number = {1},
	journaltitle = {International Journal of Computational Intelligence Systems},
	shortjournal = {{IJCIS}},
	author = {Xu, Xiaodan and Liu, Huawen and Li, Li and Yao, Minghai},
	urldate = {2025-08-25},
	date = {2018},
	langid = {english},
}

@article{fawcett_introduction_2006,
	title = {An introduction to {ROC} analysis},
	volume = {27},
	rights = {https://www.elsevier.com/tdm/userlicense/1.0/},
	issn = {01678655},
	url = {https://linkinghub.elsevier.com/retrieve/pii/S016786550500303X},
	doi = {10.1016/j.patrec.2005.10.010},
	pages = {861--874},
	number = {8},
	journaltitle = {Pattern Recognition Letters},
	shortjournal = {Pattern Recognition Letters},
	author = {Fawcett, Tom},
	urldate = {2025-08-22},
	date = {2006-06},
	langid = {english},
}

@article{byers_nearest-neighbor_1998,
	title = {Nearest-Neighbor Clutter Removal for Estimating Features in Spatial Point Processes},
	volume = {93},
	issn = {0162-1459, 1537-274X},
	url = {http://www.tandfonline.com/doi/abs/10.1080/01621459.1998.10473711},
	doi = {10.1080/01621459.1998.10473711},
	pages = {577--584},
	number = {442},
	journaltitle = {Journal of the American Statistical Association},
	shortjournal = {Journal of the American Statistical Association},
	author = {Byers, Simon and Raftery, Adrian E.},
	urldate = {2025-08-21},
	date = {1998-06},
	langid = {english},
}

@inproceedings{spence_detection_2001,
	location = {Kauai, {HI}, {USA}},
	title = {Detection, synthesis and compression in mammographic image analysis with a hierarchical image probability model},
	isbn = {978-0-7695-1336-2},
	url = {http://ieeexplore.ieee.org/document/991693/},
	doi = {10.1109/MMBIA.2001.991693},
	eventtitle = {Workshop on Mathematical Methods in Biomedical Image Analysis},
	pages = {3--10},
	booktitle = {Proceedings {IEEE} Workshop on Mathematical Methods in Biomedical Image Analysis ({MMBIA} 2001)},
	publisher = {{IEEE} Comput. Soc},
	author = {Spence, C. and Parra, L. and Sajda, P.},
	urldate = {2025-08-21},
	date = {2001},
}

@inproceedings{aleskerov_cardwatch_1997,
	location = {New York City, {NY}, {USA}},
	title = {{CARDWATCH}: a neural network based database mining system for credit card fraud detection},
	isbn = {978-0-7803-4133-3},
	url = {http://ieeexplore.ieee.org/document/618940/},
	doi = {10.1109/CIFER.1997.618940},
	shorttitle = {{CARDWATCH}},
	eventtitle = {{IEEE}/{IAFE} 1997 Computational Intelligence for Financial Engineering ({CIFEr})},
	pages = {220--226},
	booktitle = {Proceedings of the {IEEE}/{IAFE} 1997 Computational Intelligence for Financial Engineering ({CIFEr})},
	publisher = {{IEEE}},
	author = {Aleskerov, E. and Freisleben, B. and Rao, B.},
	urldate = {2025-08-21},
	date = {1997},
}

@article{edgeworth_discordant_1887,
	title = {On discordant observations},
	volume = {23},
	issn = {1941-5982, 1941-5990},
	url = {https://www.tandfonline.com/doi/full/10.1080/14786448708628471},
	doi = {10.1080/14786448708628471},
	pages = {364--375},
	number = {143},
	journaltitle = {The London, Edinburgh, and Dublin Philosophical Magazine and Journal of Science},
	shortjournal = {The London, Edinburgh, and Dublin Philosophical Magazine and Journal of Science},
	author = {Edgeworth, F.Y.},
	urldate = {2025-08-21},
	date = {1887-04},
	langid = {english},
}

@software{maus_code_2025,
	title = {Code to run simulations and reproduce results},
	url = {https://git.ae.hpi.de/philip.maus/ba-thesis-code},
	author = {Maus, Philip},
	date = {2025-08-15},
}

@misc{gemeinsamer_bundesausschuss_qualitatsberichte_2025,
	title = {Qualitätsberichte der Krankenhäuser},
	author = {{Gemeinsamer Bundesausschuss}},
	date = {2025-08-01},
}

@software{radiker_xmloot_2025,
	title = {xmloot},
	url = {https://codeberg.org/fyrk/xmloot},
	version = {1.0},
	author = {Rädiker, Flora and Maus, Philip},
	date = {2025-07-04},
}

@article{doskoc_efficient_2016,
	title = {Efficient Learning of Regular Expressions with Hierarchies},
	journaltitle = {unpublished},
	author = {Doskoč, Vanja and Friedrich, Tobias and Klodt, Nicolas and Kötzing, Timo and Naumann, Felix and Wells, Armin},
	date = {2016},
	langid = {english},
}

@article{smiti_critical_2020,
	title = {A critical overview of outlier detection methods},
	volume = {38},
	issn = {15740137},
	url = {https://linkinghub.elsevier.com/retrieve/pii/S1574013720304068},
	doi = {10.1016/j.cosrev.2020.100306},
	pages = {100306},
	journaltitle = {Computer Science Review},
	shortjournal = {Computer Science Review},
	author = {Smiti, Abir},
	urldate = {2025-06-06},
	date = {2020-11},
	langid = {english},
}

@inproceedings{pei_efficient_2006,
	location = {Hong Kong, China},
	title = {An Efficient Reference-Based Approach to Outlier Detection in Large Datasets},
	url = {http://ieeexplore.ieee.org/document/4053074/},
	doi = {10.1109/icdm.2006.17},
	eventtitle = {Sixth International Conference on Data Mining ({ICDM}'06)},
	pages = {478--487},
	booktitle = {Sixth International Conference on Data Mining ({ICDM}'06)},
	publisher = {{IEEE}},
	author = {Pei, Yaling and Zaiane, Osmar and Gao, Yong},
	urldate = {2025-07-17},
	date = {2006-12},
	note = {{ISSN}: 1550-4786},
}

@article{yang_mean-shift_2021,
	title = {Mean-shift outlier detection and filtering},
	volume = {115},
	rights = {https://www.elsevier.com/tdm/userlicense/1.0/},
	issn = {0031-3203},
	url = {https://linkinghub.elsevier.com/retrieve/pii/S0031320321000613},
	doi = {10.1016/j.patcog.2021.107874},
	pages = {107874},
	journaltitle = {Pattern Recognition},
	author = {Yang, Jiawei and Rahardja, Susanto and Fränti, Pasi},
	urldate = {2025-07-17},
	date = {2021-07},
	langid = {english},
	note = {Publisher: Elsevier {BV}},
}

@article{yuhua_li_selecting_2011,
	title = {Selecting Critical Patterns Based on Local Geometrical and Statistical Information},
	volume = {33},
	rights = {https://ieeexplore.ieee.org/Xplorehelp/downloads/license-information/{IEEE}.html},
	issn = {0162-8828, 2160-9292},
	url = {http://ieeexplore.ieee.org/document/5611541/},
	doi = {10.1109/tpami.2010.188},
	pages = {1189--1201},
	number = {6},
	journaltitle = {{IEEE} Transactions on Pattern Analysis and Machine Intelligence},
	shortjournal = {{IEEE} Trans. Pattern Anal. Mach. Intell.},
	author = {{Yuhua Li} and Maguire, L},
	urldate = {2025-07-17},
	date = {2011-06},
	note = {Publisher: Institute of Electrical and Electronics Engineers ({IEEE})},
}

@article{navarro_guided_2001,
	title = {A guided tour to approximate string matching},
	volume = {33},
	rights = {https://www.acm.org/publications/policies/copyright\_policy\#Background},
	issn = {0360-0300, 1557-7341},
	url = {https://dl.acm.org/doi/10.1145/375360.375365},
	doi = {10.1145/375360.375365},
	pages = {31--88},
	number = {1},
	journaltitle = {{ACM} Computing Surveys},
	shortjournal = {{ACM} Comput. Surv.},
	author = {Navarro, Gonzalo},
	urldate = {2025-07-14},
	date = {2001-03},
	langid = {english},
	note = {Publisher: Association for Computing Machinery ({ACM})},
}

@inproceedings{ramaswamy_efficient_2000,
	location = {Dallas Texas {USA}},
	title = {Efficient algorithms for mining outliers from large data sets},
	rights = {https://www.acm.org/publications/policies/copyright\_policy\#Background},
	url = {https://dl.acm.org/doi/10.1145/342009.335437},
	doi = {10.1145/342009.335437},
	eventtitle = {{SIGMOD}/{PODS}00: {ACM} International Conference on Management of Data and Symposium on Principles of Database Systems},
	pages = {427--438},
	booktitle = {Proceedings of the 2000 {ACM} {SIGMOD} international conference on Management of data},
	publisher = {{ACM}},
	author = {Ramaswamy, Sridhar and Rastogi, Rajeev and Shim, Kyuseok},
	urldate = {2025-07-14},
	date = {2000-05-16},
}

@article{yang_outlier_2023,
	title = {Outlier detection: How to Select k for k-nearest-neighbors-based outlier detectors},
	volume = {174},
	rights = {https://www.elsevier.com/tdm/userlicense/1.0/},
	issn = {0167-8655},
	url = {https://linkinghub.elsevier.com/retrieve/pii/S0167865523002404},
	doi = {10.1016/j.patrec.2023.08.020},
	shorttitle = {Outlier detection},
	pages = {112--117},
	journaltitle = {Pattern Recognition Letters},
	author = {Yang, Jiawei and Tan, Xu and Rahardja, Sylwan},
	urldate = {2025-07-11},
	date = {2023-10},
	langid = {english},
	note = {Publisher: Elsevier {BV}},
}

@misc{deutsches_institut_fur_normung_din_2020,
	title = {{DIN} {ISO} 8601-1:2020-12, Datum und Uhrzeit\_- Darstellung für den Informationsaustausch\_- Teil\_1: Grundlegende Regeln ({ISO}\_8601-1:2019)},
	url = {https://dx.doi.org/10.31030/3178552},
	doi = {10.31030/3178552},
	shorttitle = {{DIN} {ISO} 8601-1},
	publisher = {{DIN} Media {GmbH}},
	author = {{Deutsches Institut für Normung}},
	urldate = {2025-06-16},
	date = {2020-12},
}

@article{campos_evaluation_2016,
	title = {On the evaluation of unsupervised outlier detection: measures, datasets, and an empirical study},
	volume = {30},
	issn = {1384-5810, 1573-756X},
	url = {http://link.springer.com/10.1007/s10618-015-0444-8},
	doi = {10.1007/s10618-015-0444-8},
	shorttitle = {On the evaluation of unsupervised outlier detection},
	pages = {891--927},
	number = {4},
	journaltitle = {Data Mining and Knowledge Discovery},
	shortjournal = {Data Min Knowl Disc},
	author = {Campos, Guilherme O. and Zimek, Arthur and Sander, Jörg and Campello, Ricardo J. G. B. and Micenková, Barbora and Schubert, Erich and Assent, Ira and Houle, Michael E.},
	urldate = {2025-07-07},
	date = {2016-07},
	langid = {english},
}

@inproceedings{breunig_lof_2000,
	location = {Dallas Texas {USA}},
	title = {{LOF}: identifying density-based local outliers},
	isbn = {978-1-58113-217-5},
	url = {https://dl.acm.org/doi/10.1145/342009.335388},
	doi = {10.1145/342009.335388},
	shorttitle = {{LOF}},
	eventtitle = {{SIGMOD}/{PODS}00: {ACM} International Conference on Management of Data and Symposium on Principles of Database Systems},
	pages = {93--104},
	booktitle = {Proceedings of the 2000 {ACM} {SIGMOD} international conference on Management of data},
	publisher = {{ACM}},
	author = {Breunig, Markus M. and Kriegel, Hans-Peter and Ng, Raymond T. and Sander, Jörg},
	urldate = {2025-07-07},
	date = {2000-05-16},
	langid = {english},
}

@article{hodge_survey_2004,
	title = {A Survey of Outlier Detection Methodologies},
	volume = {22},
	rights = {https://www.springernature.com/gp/researchers/text-and-data-mining},
	issn = {0269-2821, 1573-7462},
	url = {https://link.springer.com/10.1023/B:AIRE.0000045502.10941.a9},
	doi = {10.1023/b:aire.0000045502.10941.a9},
	pages = {85--126},
	number = {2},
	journaltitle = {Artificial Intelligence Review},
	author = {Hodge, Victoria and Austin, Jim},
	urldate = {2025-05-23},
	date = {2004-10},
	langid = {english},
	note = {Publisher: Springer Science and Business Media {LLC}},
}

@article{chandola_anomaly_2009,
	title = {Anomaly detection: A survey},
	volume = {41},
	issn = {0360-0300, 1557-7341},
	url = {https://dl.acm.org/doi/10.1145/1541880.1541882},
	doi = {10.1145/1541880.1541882},
	shorttitle = {Anomaly detection},
	pages = {1--58},
	number = {3},
	journaltitle = {{ACM} Computing Surveys},
	shortjournal = {{ACM} Comput. Surv.},
	author = {Chandola, Varun and Banerjee, Arindam and Kumar, Vipin},
	urldate = {2025-05-23},
	date = {2009-07},
	langid = {english},
}

@Preamble
{
    {
    \newcommand{\bibciac}[2]{Proceedings of the #1 Conference on Algorithms and Complexity (CIAC'#2)}
    \newcommand{\bibdac}[2]{Proceedings of the #1 Annual Design Automation Conference (DAC'#2)}
    \newcommand{\bibinvisau}[1]{Proceedings of the Australian Symposium on Information Visualisation (invis.au #1)}
    \newcommand{\bibieeepdp}[2]{Proceedings of the #1 IEEE Symposium on Parallel and Distributed Processing #2}
    \newcommand{\bibieeecs}[1]{Proceedings of the IEEE International Symposium on Circuits and Systems #1}
    \newcommand{\bibcccg}[2]{Proceedings of the #1 Canadian Conference on Computational Geometry (CCCG'#2)}
    \newcommand{\bibswat}[2]{Proceedings of the #1 Scandinavian Workshop on Algorithm Theory (SWAT'#2)}
    \newcommand{\bibipco}[2]{Proceedings of the #1 International Conference on Integer Programming and Combinatorial Optimization (IPCO'#2)}
    \newcommand{\bibsofsem}[2]{Proceedings of the #1 Conference on Current Trends in Theory and Practice of Computer Science (SOFSEM'#2)}
    \newcommand{\bibstoc}[2]{Proceedings of the #1 Annual ACM Symposium on Theory of Computing (STOC'#2)}
    \newcommand{\bibfocs}[2]{Proceedings of the #1 Annual Symposium on Foundations of Computer Science (FOCS'#2)}
    \newcommand{\bibsoda}[2]{Proceedings of the #1 Annual ACM-SIAM Symposium on Discrete Algorithms (SODA'#2)}
    \newcommand{\bibgd}[2]{Proceedings of the #1 International Symposium on Graph Drawing (GD'#2)}
    \newcommand{\bibinfovis}[1]{Proceedings of the IEEE Symposium on Information Visualization (InfoVis'#1)}
    \newcommand{\bibvis}[1]{Proceedings of the IEEE Conference on Visualization (Vis'#1)}
    \newcommand{\bibpvis}[1]{Proceedings of the IEEE Pacific Visualisation Symposium (PacificVis'#1)}
    \newcommand{\bibsoftvis}[2]{Proceedings of the #1 ACM Symposium on Software Visualization (SoftVis'#2)}
    \newcommand{\bibeurocg}[2]{Proceedings of the #1 European Workshop on Computational Geometry (EuroCG'#2)}
    \newcommand{\bibsocg}[2]{Proceedings of the #1 Annual Symposium on Computational Geometry (SoCG'#2)}
    \newcommand{\bibwads}[2]{Proceedings of the #1 International Symposium on Algorithms and Data Structures (WADS'#2)}
    \newcommand{\bibwg}[2]{Proceedings of the #1 Workshop on Graph-Theoretic Concepts in Computer Science (WG'#2)}
    \newcommand{\bibgta}{Proceedings of the Conference at Graph Theory and Applications}
    \newcommand{\bibisaac}[2]{Proceedings of the #1 International Symposium on Algorithms and Computation (ISAAC'#2)}
    \newcommand{\bibcocoon}[2]{Proceedings of the #1 Annual International Conference on Computing and Combinatorics (COCOON'#2)}
    \newcommand{\bibtamc}[2]{Proceedings of the #1 Annual Conference on Theory and Applications of Models of Computation (TAMC'#2)}
    \newcommand{\bibicalp}[2]{Proceedings of the #1 International Colloquium on Automata, Languages and Programming (ICALP'#2)}
    \newcommand{\biblatin}[2]{Proceedings of the #1 Latin American Symposium (LATIN'#2)}
    \newcommand{\bibesa}[2]{Proceedings of the #1 Annual European Symposium on Algorithms (ESA'#2)}
    }
}

@String{IEEE
= {IEEE Computer Society}}

@String{Elsevier
= {Elsevier Science Publishers}}

@String{ACM
= {ACM Press}}

    \addchap{Declaration of Authorship}
    I hereby declare that this thesis is my own unaided work. All direct or indirect sources used are acknowledged as references.\\[6 ex]

\begin{flushleft}
    Potsdam, \today
    \hspace*{2 em}
    \raisebox{-0.9\baselineskip}
    {
        \begin{tabular}{p{5 cm}}
            \hline
            \centering\footnotesize\printAuthor
        \end{tabular}
    }
\end{flushleft}

\end{document}